\documentclass[10pt,twocolumn,letterpaper]{article}

\usepackage[pagenumbers]{cvpr} % To force page numbers, e.g. for an arXiv version

\usepackage{graphicx}
\usepackage{amsmath}
\usepackage{amssymb}
\usepackage{booktabs}
\usepackage{multirow}
\usepackage{nicefrac}       % compact symbols for 1/2, etc.
\usepackage{microtype}      % microtypography
\usepackage[table,xcdraw]{xcolor}
\usepackage{comment}
\usepackage{grffile}
\usepackage{hyperref}
\hypersetup{pagebackref,breaklinks,colorlinks}

\usepackage[capitalize]{cleveref}
\crefname{section}{Sec.}{Secs.}
\Crefname{section}{Section}{Sections}
\Crefname{table}{Table}{Tables}
\crefname{table}{Tab.}{Tabs.}

\definecolor{gold}{RGB}{221, 196, 65}
\definecolor{silver}{RGB}{215, 215, 215}
\definecolor{bronze}{RGB}{126, 66, 5}

\def\cvprPaperID{10771} % *** Enter the CVPR Paper ID here
\def\confName{CVPR}
\def\confYear{2023}

\begin{document}

%%%%%%%%% TITLE - PLEASE UPDATE
\title{4K-NeRF: High Fidelity Neural Radiance Fields at Ultra High Resolutions}
% NeRF-UHD
\author{
Zhongshu Wang,
Lingzhi Li,
Zhen Shen,
Li Shen,
Liefeng Bo \\
Alibaba Group\\
Beijing, China\\
{\tt\small \{zhongshu.wzs, llz273714, zackary.sz, jinyan.sl, liefeng.bo\}@alibaba-inc.com}
% For a paper whose authors are all at the same institution,
% omit the following lines up until the closing ``}''.
% Additional authors and addresses can be added with ``\and'',
% just like the second author.
% To save space, use either the email address or home page, not both
% \and
% Lingzhi Li\\
% Alibaba Group\\
% Beijing, China\\
% {\tt\small llz273714@alibaba-inc.com}
% Institution2\\
% First line of institution2 address\\
% {\tt\small secondauthor@i2.org}
}

\twocolumn[{%
\renewcommand\twocolumn[1][]{#1}%
\vspace{-1em}
\maketitle
\vspace{-1em}
\begin{center}
\centering

\includegraphics[width=1.0\linewidth]{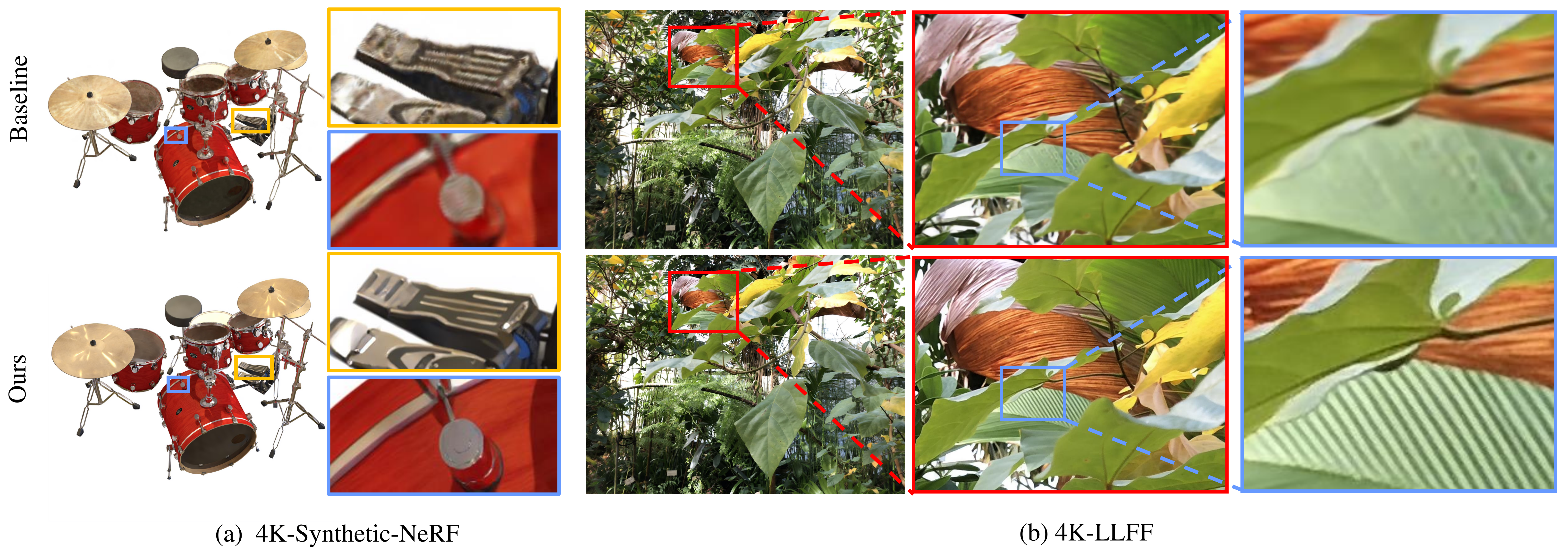}
\captionof{figure}{Visual comparison between DVGO and our method on example scenes in 4K-Synthetic-NeRF (a) and 4K-LLFF (b).
}
\label{fig:teaser}
\end{center}%
}]

% \begin{figure*}[tp]
%    \centering
%    %\fbox{\rule{0pt}{2in} \rule{0.9\linewidth}{0pt}}
%    \includegraphics[width=1\linewidth]{figure/teaser.png}

%    \caption{\textbf{Qualitative comparison between our method (left) and DVGO (right).} The picture is better displayed in a high resolution display for visualization.}
%    \label{fig:cp2}
%    \vspace{-0.5em}
% \end{figure*}

%%%%%%%%% ABSTRACT
\begin{abstract}
In this paper, we present a novel and effective framework, named 4K-NeRF, to pursue high fidelity view synthesis on the challenging scenarios of ultra high resolutions, building on the methodology of neural radiance fields (NeRF). The rendering procedure of NeRF-based methods typically relies on a pixel-wise manner in which rays (or pixels) are treated independently on both training and inference phases, limiting its representational ability on describing subtle details, especially when lifting to a extremely high resolution. We address the issue by exploring ray correlation to enhance high-frequency details recovery. Particularly, we use the 3D-aware encoder to model geometric information effectively in a lower resolution space and recover fine details through the 3D-aware decoder, conditioned on ray features and depths estimated by the encoder. Joint training with patch-based sampling further facilitates our method incorporating the supervision from perception oriented regularization beyond pixel-wise loss. Benefiting from the use of geometry-aware local context, our method can significantly boost rendering quality on high-frequency details compared with modern NeRF methods, and achieve the state-of-the-art visual quality on 4K ultra-high-resolution scenarios. Code Available at \url{https://github.com/frozoul/4K-NeRF}
\end{abstract}

%  NeRF is one of the most emergency Novel View Synthesis technologies, capable of synthesis visually realistic new perspective images. However, all existing NeRF methods only work well at low resolutions (less or equal to 1K). Visual artifacts and rendering costs increase significantly once the rendering resolution grows. In order to solve the problem of NeRF in ultra-high-resolution images, we propose a novel NeRF method. By improving the ray sampling strategy, we introduce the convolutional network into NeRF for the first time to take advantage of the local pattern of images. On the other hand, we use a conditional depth map to guide convolutional network training, ensuring the view consistency under high-resolution images and improving the hierarchical sense of details. Finally, the convolutional network can use more diversified loss functions to achieve a better distortion and visual perception tradeoff. Experiments show that our method achieves the SOTA visual quality and the fastest rendering speed in ultra-high resolution.

%%%%%%%%% BODY TEXT
\section{Introduction}\label{sec:intro}

\begin{figure*}[th]
  \centering
   \includegraphics[width=0.98\linewidth]{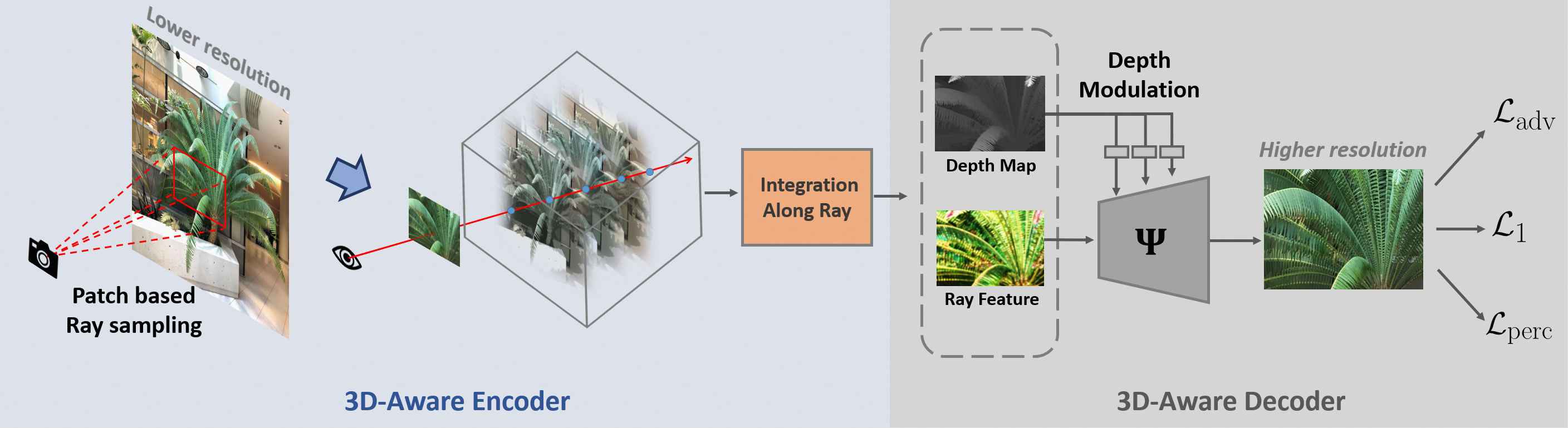}
   \caption{\textbf{The overall pipeline of 4K-NeRF.} Using patch-based ray sampling, we jointly train the 3D-Aware encoder for embedding 3D geometric information in a lower resolution space and the 3D-Aware decoder for realizing rendering  enhancement on high-frequency details in the full resolution space.} 
   %Furthermore, Geometric information is shared between the two modules through multiple layers of modulation and gradient propagation. Finally, we use GAN loss to promote the closeness of the overall feature distribution of the rendering images and ground truth.}
   \label{fig:pipe}
\end{figure*}

Ultra-High-Resolution has growing popular as a standard for recording and displaying images and videos, even supported in modern mobile devices. A scene captured in ultra high resolution format typically presents content with incredible details compared to using a relatively lower resolution (e.g, 1K high-definition format) in which the information at a pixel is enlarged by a small patch in extremely high resolution images. Developing techniques for handling such high-frequency details poses challenges for a wide range of tasks in image processing and computer vision. In this paper, we focus on the novel view synthesis task and investigate the potential of realizing high fidelity view synthesis rich in subtle details at ultra high resolution. 

Novel view synthesis aims to produce free-view photo-realistic synthesis captured for a scene from a set of viewpoints. Recently, Neural Radiance Fields \cite{NeRF} offer a new  methodology for modeling and rendering 3D scenes by virtue of deep neural networks and have demonstrated remarkable success on improving visual quality compared to traditional view interpolation methods \cite{YaoYao2018MVSNetDI,tucker2020single}. Particularly, a mapping function, instantiated as a deep multilayer perceptron (MLP), is optimized to associate each 3D location given a viewing direction to its corresponding radiance color and volume density, while realizing view-dependent effect requires querying the large network hundreds of times for casting a ray through each pixel.
Several following approaches are proposed to improve the method either from the respect of reducing aliasing artifacts on multiple scales \cite{mipnerf} or improving training and inference efficiency benefiting from the use of discretized structures \cite{yu2021plenoxels, ChengSun2022DirectVG, TensoRF}. All these methods follow the pixel-wise mechanism despite varying architectures, i.e., rays (or pixels) are regarded individually during training and inference phase. They are typically developed on training views up to 1K resolution. When applying the approaches on ultra-high-resolution scenarios, they would struggle with objectionable blurring artifacts  (as shown in Fig.~\ref{fig:4kall}) due to insufficient representational ability for capturing fine details.

In this paper, we introduce a novel framework, named 4K-NeRF, building upon the methodology of NeRF-based volume rendering to realize high fidelity view synthesis at 4K ultra-high-resolution. We take the inspiration from the success of convolutional neural networks on traditional super resolution \cite{dong2015image}.
%which serves to resolve a lower resolution observation to a higher resolution with rich details by the aid of local priors captured between neighbouring pixels. 
We expect to boost the representational power of NeRF-based methods by better exploring local correlations between rays.

Specially, the framework is comprised of two components, a 3D-aware encoder and a 3D-aware decoder, as shown in Fig.~\ref{fig:pipe}. The encoder encodes geometric properties of a scene effectively in a lower resolution space, forming intermediate ray features and geometry information (i.e., estimated depth) feeding into the decoder. The decoder is capable of recovering high-frequency details by integrating geometry-aware local patterns learned through depth-modulated convolutions in the higher resolution (full-scale) observations. We further introduce a patch-based ray sampling strategy replacing the random sampling in NeRFs, allowing the encoder and the decoder trained jointly with the perception-oriented losses complementing to the conventional pixel-wise MSE loss. Such a joint training facilitates coordinating geometric modelling in the encoder with local context learning in the decoder.  Compared to traditional pixel-wise mechanism \cite{NeRF, ChengSun2022DirectVG} our method can realize significant enhancement on fine details even with an extreme zoom-in extent (as shown in Fig.~\ref{fig:teaser}). Extensive comparison and ablation studies on synthetic and real-world scenes demonstrate the effectiveness of the proposed framework both quantitatively and qualitatively.  
%------------------------------------------------------------------------
\section{Related work}
\label{sec:formatting}
\textbf{Neural Radiance Fields.}
NeRF \cite{NeRF} formulate a continuous mapping from coordinates to corresponding color and density through MLP via differentiable volume rendering and achieves impressive rendering quality. Some approaches realize significant acceleration on training and rendering speed by virtue of explicit structures, e.g., octree-based structure \cite{yu2021plenoctrees_iccv}, dense \cite{ChengSun2022DirectVG} or sparse voxel grids \cite{SNeRG, nsvf, kilonerf, yu2021plenoxels} or a hybrid structure \cite{hu2022efficientnerf}, radiance maps \cite{fastnerf}, and tri-planes \cite{TensoRF}.
Some methods focus on improving the rendering quality of NeRF from different aspects. The works in \cite{mipnerf, mipnerf360} leverage the insight of mipmap to achieve anti-aliasing. \cite{verbin2021refnerf, Nex} improve representational ability on modelling specular reflections. A series of methods are developed on sparse views, either benefiting from depth prior \cite{deng2022depth,roessle2022dense} or taking as input image features extracted from 2D convolutional networks \cite{yu2021pixelnerf, AnpeiChen2021MVSNeRFFG}. 
To the best of our knowledge,  our framework is the first to successfully extend NeRF-based paradigm to 4K resolution, proving high-fidelity viewing experience with crystal-clear and high-frequency details.

\textbf{Novel View Synthesis.} 
Apart from directly approximate a radiance fields for image synthesis, many effort have been done by the research community for view synthesis, mainly representing the 3D scene with the data structure of mesh \cite{shih20203d,hu2021worldsheet}, %3D geometry-based \cite{PaulDebevec1996ModelingAR,NoahSnavely2006PhotoTE,PeterHedman2017Casual3P}, 
point cloud~\cite{wiles2020synsin} and multiplane images~\cite{tucker2020single,JohnFlynn2019DeepViewVS,MichaelBroxton2020ImmersiveLF}. Some recent methods use CNN or transformer to enhance visual quality.
%PixelNeRF\llz{cite} extracts scene prior with a learned CNN to handle very sparse input,  
IBRNet \cite{IBRNetLM} enables large-scale reasoning with a ray transformer. DeepVoxel \cite{DeepVoxelsLP} takes advantage of both 3D and 2D CNNs to achieve better 3D representation and improve final render quality. EG3D \cite{eg3d} applies a 2D upsample module to increase the resolution of generated faces.
%The advantage of these methods is rendering speed, but their results were unsatisfactory. 
GIRAFFE \cite{MichaelNiemeyer2020GIRAFFERS} realizes a compositional generative model incorporating with 2D neural rendering via 2D convolutional networks. In contrast, our method realize extremely high-resolution view synthesis by better exploiting geometry-aware local patterns, i.e., enhancing correlation of ray features via depth-modulated convolutions.

\textbf{High-Resolution Synthesis.} The framework is also related to image super-resolution techniques which recover high-resolution images from low-resolution ones. Classical methods are typically derived from strong prior on ideal image degradation type, i.e., downsampling and noisy \cite{dong2015image, ji2020real}. These methods investigate gradient propagation in low-level network layers \cite{zhang2018image,dai2019second} or the balance between distortion and perception \cite{ledig2017photo,wang2018esrgan}. In order to address more complex scenarios, some methods introduce first-order \cite{zhang2021designing,liang2021swinir} and high-order hybrid degradation modeling \cite{wang2021real}, and have achieved promising performance on real-world data.
All of these super-resolution methods perform on resolving 2D single image. The most related work to ours is NeRF-SR \cite{nerfsr}, which incorporates super-resolution/sampling into NeRFs. Unlike using a joint training scheme in our framework, the method trains a separate refinement network to super-resolve image patches by using the max-pooled features of relevant patches sampled from higher-resolution references, resulting in less-consistent rendering across viewpoints. 
\section{Method}
%We first go over the methodology of NeRF-based volumetric rendering and discuss the limitation on modeling and rendering the scenes with extremely high resolution. We then present our NeRF-4K framework in detail and introduce the training strategy with loss functions in the next section.

\subsection{Volumetric Rendering}
NeRF realizes view synthesis by learning a continuous mapping function to estimate the color $\mathbf{c}\in\mathbb{R}^3$ and the volume density $\sigma\in\mathbb{R}$ of a 3D point position $\mathbf{x}\in \mathbb{R}^3$ and a viewing direction $\mathbf{d} \in \mathbb{R}^3$, i.e., $ \Phi: (\mathbf{x},\mathbf{d})\mapsto  (\mathbf{c}, \sigma)$. To render an image given camera pose, the expected color $\widehat{\mathbf{C}}(\mathbf{r})$ of a camera ray $\mathbf{r} = \mathbf{o} + t\mathbf{d}$ 
%from the camera center $\mathbf{o}$ 
through the pixel is estimated by sampling a set of points along the ray and integrating their colors to approximate a volumetric rendering integral \cite{OpticalModel},
% with the the numerical quadrature \cite{OpticalModel} discussed by Max, 
\begin{equation}\label{eq:render} 
    \widehat{\mathbf{C}}(\mathbf{r}) = \sum_{i=1}^N T_i \cdot \alpha_i \cdot \mathbf{c}_i,
\end{equation}
\begin{equation}\label{eq:alpha_t}
    \alpha_i = 1 - \exp(-\sigma_i \delta_i),\quad T_i = \prod_{j=1}^{i-1} (1 - \alpha_j),
\end{equation}
where $\alpha_i$ denotes the ray termination probability at the point $i$, $\delta_i = t_{i+1} - t_i$ represents the distance between two adjacent points, and $T_i$ indicates the
accumulated transmittance when reaching $i$. The mapping function $\Phi$ is instantiated as a MLP. Given the training views with known poses, the model is trained by minimizing the mean squared errors (MSE) between the predicted pixel colors and the ground-truth colors,
\begin{equation}\label{eq:mse_loss}
    \mathcal{L}_{\text{MSE}} = \frac{1}{|\mathcal{R}|}\sum_{\mathbf{r}\in \mathcal{R}}\left\|\widehat{\mathbf{C}}(\mathbf{r}) - \mathbf{C}(\mathbf{r})\right\|_2^2,
\end{equation}
where $\mathcal{R}$ denotes the ray set randomly sampled in each min-batch. The optimization of each point is according to its projection through the rays of different viewpoints.
Some variants are distinct from using single large neural networks, by integrating the benefit of explicit structures \cite{kilonerf, nsvf, yu2021plenoxels, ChengSun2022DirectVG}, while 
%By confining density prediction to the use of positions, 
all these methods intrinsically learn geometry-aware representations in pixel-wise manner despite architecture difference. 

\textbf{Limitation.} Rays (or pixels) are treated independently during training and inference process. The cardinality of the ray set grows quadratically with the increase of image resolution. For an image of 4K ultra high resolution, there exists over 8 million pixels typically presenting richer details and each of which naturally embodies scene content in a finer level than the one on a lower resolution image. If directly using such a pixel-wise training mechanism on extremely high-resolution inputs, these methods may struggle with insufficient representational ability for retaining subtle details, even with increased model capacity (shown in the supplementary materials), which might worsen the issues of lengthy inference with a tremendous MLP or considerable storage cost by using voxel-grid structures with increased volume dimension.

\subsection{Overall Framework}
To extend conventional NeRF methods to achieve high-quality rendering at ultra high resolutions, one straightforward solution is to first train NeRF models for rendering down-sampled outputs and then train parameterized super-resolution on each view to up-sample them to full scale. However, such a solution would result in obvious artifacts of inconsistent rendering across viewpoint, as local patterns captured in the super-resolution stage lack regularization from holistic geometry (as shown in the ablation study of joint training in~\ref{as: jointrain}).

In this regard, we develop a simple yet effective framework which first encodes geometric information in a lower resolution space through {\it 3D-Aware Encoder}  module and recover subtle details in a higher resolution (HR) space via the {\it 3D-Aware Decoder} module. 
The method aims to boost high-frequency details recovery by integrating 3D-aware local correlations learned in the observations.

\subsection{3D-Aware Encoder}
We instantiate the encoder based on the formulation defined in the DVGO \cite{ChengSun2022DirectVG}, where voxel-grid based representations are learned to encode geometric structure explicitly,
\begin{equation}
    \left ( \textbf{x}, \textbf{V} \right ):\left ( \mathbb{R}^{3},\mathbb{R}^{N_c\times N_{x}\times N_{y}\times N_{z}}\right )\rightarrow \mathbb{R}^{N_c},
\end{equation}
where $N_c$ denotes the channel dimension for density ($N_c = 1$) and color modality, respectively. For each sampling point, the density is estimated by trilinear interpolation equipped with a softplus activation, i.e., $\sigma= \text{softplus}\left (\text{interp}\left (\textbf{x},\textbf{V}_d \right )\right )$. The colors are estimated with a shallow MLP,
\begin{equation}
\begin{aligned}
    \mathbf{c} &= f_{\text{MLP}}\left (\text{interp}\left ( \textbf{x},\textbf{V}_c \right ),\textbf{x},\textbf{d} \right ) \\
    &= f_{\text{RGB}}\big(g_{\theta}(\text{interp}(\textbf{x},\textbf{V}_c),\textbf{x},\textbf{d})\big),
\end{aligned}
\end{equation}
where $g_\theta(\cdot)$ extracts volumetric features for color information, and $f_\text{RGB}$ denotes the mapping (with one or multiple layers) from the features to RGB images. 

The output $\mathbf{g} = g(\theta; \mathbf{x}, \mathbf{d})$ denotes the volumetric feature for the point $\mathbf{x}$ with the viewing direction $\mathbf{d}$. We can then get the descriptor for each ray (or pixel) by accumulating the features of sampling points along the ray $\mathbf{r}$ as in Eqn.\ref{eq:render} ,
\begin{equation}\label{eq: ray_feature} 
    \mathbf{f}(\mathbf{r}) = \sum_{i=1}^N T_i \cdot \alpha_i \cdot \mathbf{g}_i.
\end{equation}
For better use of geometric properties embedded in the encoder, we also generate a depth map by estimating the depth along the camera axis for each ray $\mathbf{r}$,
\begin{equation}\label{eq: depth} 
    M(\mathbf{r}) = \sum_{i=1}^N T_i \cdot \alpha_i \cdot t_i, 
\end{equation}
where $t_i$ denotes the distance of the sampling point $i$ to the camera center as in Eqn.\ref{eq:render}. The estimated depth map provides a strong guidance for understating the 3D structure of a scene, e.g., nearby pixels on the image plane may be far away in the original 3D space. 
Assume the spatial dimension is $H'\times W'$, the formed feature maps $\mathbf{F}_{\text{en}} \in \mathbb{R}^{C'\times H'\times W'}$ and the depth map $\mathbf{M}\in \mathbb{R}^{H' \times W'}$ are fed into the decoder for pursuing high-fidelity reconstruction of fine details.
% balance geometric 
\subsection{3D-Aware Decoder}
%by taking the feature maps $\mathbf{F}_{\text{en}}\in \mathbb{R}^{C'\times H'\times W'}$ and $\mathbf{M}\in \mathbb{R}^{H'\times W'}$ as input and 
The decoder performs view synthesis at a higher spatial dimension $H\times W$ space by training a convolutional neural network $\Psi: \left(\mathbf{F_\text{en}},\mathbf{M}\right)\mapsto \mathbf{P}$, where $\mathbf{P}\in \mathbb{R}^{3\times H\times W}$, $H= sH'$ and $W = sW'$, and $s$ indicates the up-sampling scale. The network is built by stacking several convolutional blocks (with neither non-parametric normalization nor down-sampling operations) interleaved with up-sampling operations. Particularly, instead of simply concatenating the features $\mathbf{F}_{\text{en}}$ and the depth map $\mathbf{M}$, we regard depth signal separately and inject it into every block through a learned transformation to modulate block activation.  
% \li{the architecture can be mentioned in the implementation section} 

Formally, suppose $\mathbf{F}^k$ denotes the activation of an intermediate block with the channel dimension $C_k$. The depth map $\mathbf{M}$ passes through the transformation (e.g., with $1\times 1$ convolution) to predict scale and bias values with the same dimension $C_k$, used to modulate $\mathbf{F}^k$ according to:
\begin{equation}\label{eq:scale_bias}
\tilde{\mathbf{F}}^k_{i,j} = \gamma_{i,j}^k(\mathbf{M}) \odot \mathbf{F}^k_{i,j} + \beta_{i,j}(\mathbf{M}).
\end{equation}
where $\odot$ denotes element-wise product, $i$ and $j$ indicate the spatial position. More detailed descriptions for the network architecture can be founded in the implementation section and supplemental material.

Integrating local information of nearby pixels has proven to be effective for recovering high frequency details in single image super-resolution. 
%Pursing high quality super-resolution with view consistency is necessary for the decoder. 
Learning local correlation of ray features naturally connects pattern extraction across spacial regions to the underlying 3D geometric structure, and the modulation with depth maps further regularize the learning with geometric guidance.

\section{Training}
The encoder and the decoder are jointly trained and the overall framework can be trained in a differentiable and end-to-end manner. 

\textbf{Patch-based Ray Sampling.}  Our method aims to capture spatial information between rays (pixels). Therefore, the random ray sampling strategy used in traditional NeRF methods is unsuitable here. We present a training strategy with patch-based ray sampling to facilitate the capture of spatial dependencies between ray features.

We first split the images of training views into patches $\mathbf{p}$ with the size $N_p\times N_p$ in order to ensure the sampling probability on pixels are uniform. When the image spatial dimension can not be exactly divided by the patch size, we truncate the patch until edge and obtain a set of training patches. A patch (or multiple patches) is randomly sampled from the set, and the rays casting through the pixels in the patch form the mini-batch of each iteration.  

\textbf{Loss Functions.} We found that only using distortion-oriented loss (e.g., MSE, $\ell_1$ and Huber loss) 
as objective tends to produce blurry or over-smoothed visual effects on fine details. In order to solve the problem, we add the adversarial loss and the perceptual loss to regularize fine detail synthesis. 
% The adversarial loss $\mathcal{L}_{\text{adv}}$ is defined by following the formulation \cite{}, 
The adversarial loss $\mathcal{L}_{\text{adv}}$ is calculated on the predicted image patches via the decoder and training patches through a learnable discriminator which aims to distinguish the distribution of training data and predicted one.
The perceptual loss $\mathcal{L}_{\text{perc}}$ estimates the similarity between predicted patches $\hat{\mathbf{p}}$ and Ground-Truth $\mathbf{p}$ in the feature space via a pretrained 19-layer VGG network $\varphi$ \cite{Simonyan15},
\begin{equation}\label{eq:perception}
    \mathcal{L}_{\text{perc}} = \left\| \varphi(\hat{\mathbf{p}}) - \varphi(\mathbf{p})\right\|_2^2.
\end{equation}

We use $\ell_1$ loss instead of MSE for supervising the reconstruction of high-frequency details,
\begin{equation}\label{eq:l1}
 \mathcal{L}_1 = \frac{1}{N_p^2}\left|\mathbf{C}({\hat{\mathbf{p}})} - \mathbf{C}(\mathbf{p})\right|.
\end{equation}
We add an auxiliary MSE loss to facilitate the training of encoder with down-scaled training views, i.e., the ray features produced by the encoder are fed into an extra fully-connected layer to regress RGB values in the lower-resolution images. 
The overall training objective is defined as,
\begin{equation}\label{eq:loss}
    \mathcal{L} = \lambda_h \mathcal{L}_{\text{1}} + \lambda_a \mathcal{L}_{\text{adv}} + \lambda_p \mathcal{L}_{\text{perc}} + \lambda_l \mathcal{L}_{\text{MSE}}^{l}.
\end{equation}
where $\lambda_h$, $ \lambda_a$, $ \lambda_p$ and $ \lambda_l$ denote the hyper-parameters for weighting the losses.

%\textbf{Discussion.} The design of learning features in multiple scales may be related to the coarse-to-fine training in NeRF, while there is intrinsic difference in motivation and objective. NeRF use the ``coarse" network to  

\section{Experiments}

\subsection{Implementation}
\textit{3D-Aware Encoder.}
We use the configuration of DVGO as the default setting for the encoder. Specially, we extract the ray features at the penultimate layer of the MLP (with the channel dimension $64$) following a dimensional reduction layer (with the channel dimension $6$), then the obtained features are fed into the decoder.
%as the encoder training setting in our experiments. Specifically, the size of the voxel is 384x384x256, and each voxel contains a density value representing geometry and a 12-dimensional color feature. Before color features feed into the decoder, we first use 1 MLP layer with 64 channels to transform color features. Then we use a layer of MLP to reduce 64-dimensional features to 6-dimensional conducive to optimizing the decoder. So our volume rendering operates in a 6-dimensional feature space rather than a 3-dimensional RGB space. To improve the convergence speed, we first train DVGO with low-resolution images (1008x756) to initialize the encoder.

\textit{3D-Aware Decoder.}
We employ a residual skip-connected convolutional blocks \cite{wang2018esrgan} for the decoder. Specifically, the decoder consists of a backbone with 5 blocks and an up-sampling head to produce full-scale images. We plug the depth modulation module at the end of each block. Detailed architecture can refer to supplemental material.

%\li{add in supplemental materials. Each block consists of 15 convolutional layers and three modulation modules. There is a set of dense connections between every five convolutional layers to enhance the complementarity of information extracted from each other.} There is a modulation module every five convolutional layers to introduce geometry distribution modulation based on depth information. There is also a residual connection between the three modulation modules to facilitate the propagation of gradients. This method of extracting information in the original resolution and then increasing the resolution at the end reduces much computation compared to increasing the resolution at the beginning. On the other hand, it will also have a better effect.

\textit{Training.}
To facilitate training convergence, in practice we initialize the encoder by pretraining it with 30k iterations following the training setting of DVGO . We then jointly train the encoder and the decoder for 200k iterations with patch size of 64. The loss parameters $\lambda_{h}$, $\lambda_{p}$, $\lambda_{a}$ and $\lambda_{l}$ are respectively set to 1.0, 0.5, 0.02 and 1.0. The learning rates for updating the encoder and the decoder are 1e-4 and 2e-4.
%Our training strategy has two phases. The first phase uses random ray sampling to train a DVGO with low-resolution images. 
%The second phase uses the pre-trained DVGO to initialize the VC-encoder and patch-based ray sampling to joint-train the decoder and encoder. In addition to the encoder's loss based on low-resolution ground truth, the decoder's loss based on high-resolution ground truth will be added, including $\lambda_{h}$ $\mathcal{L}_{\text{MSE}}^{h}$, $\lambda_{p}$ $\mathcal{L}_{\text{perc}}$, and $\lambda_{a}$ $\mathcal{L}_{\text{adv}}$. In the experiment, $\lambda_{h}$ is 1, $\lambda_{p}$ is 0.5, and $\lambda_{a}$ is 0.02. 
%The learning rate of the VC-Encoder is 1e-4, and the learning rate of the VC-Decoder is 2e-4. 
%The first stage has 30000 epochs, and the second stage is trained for 200000 epochs.

% \begin{figure*}
%   \centering
%   \begin{subfigure}{0.4\linewidth}
%     \fbox{\rule{0pt}{2in} \rule{.9\linewidth}{0pt}}
%     \caption{An example of a subfigure.}
%     \label{fig:short-a}
%   \end{subfigure}
%   \hfill
%   \begin{subfigure}{0.4\linewidth}
%     \fbox{\rule{0pt}{2in} \rule{.9\linewidth}{0pt}}
%     \caption{Another example of a subfigure.}
%     \label{fig:short-b}
%   \end{subfigure}
%   \caption{Comparison of view consistency effect in different setting.}
%   \label{fig:short}
% \end{figure*}

\begin{figure*}
  \centering
   \includegraphics[width=0.93\linewidth]{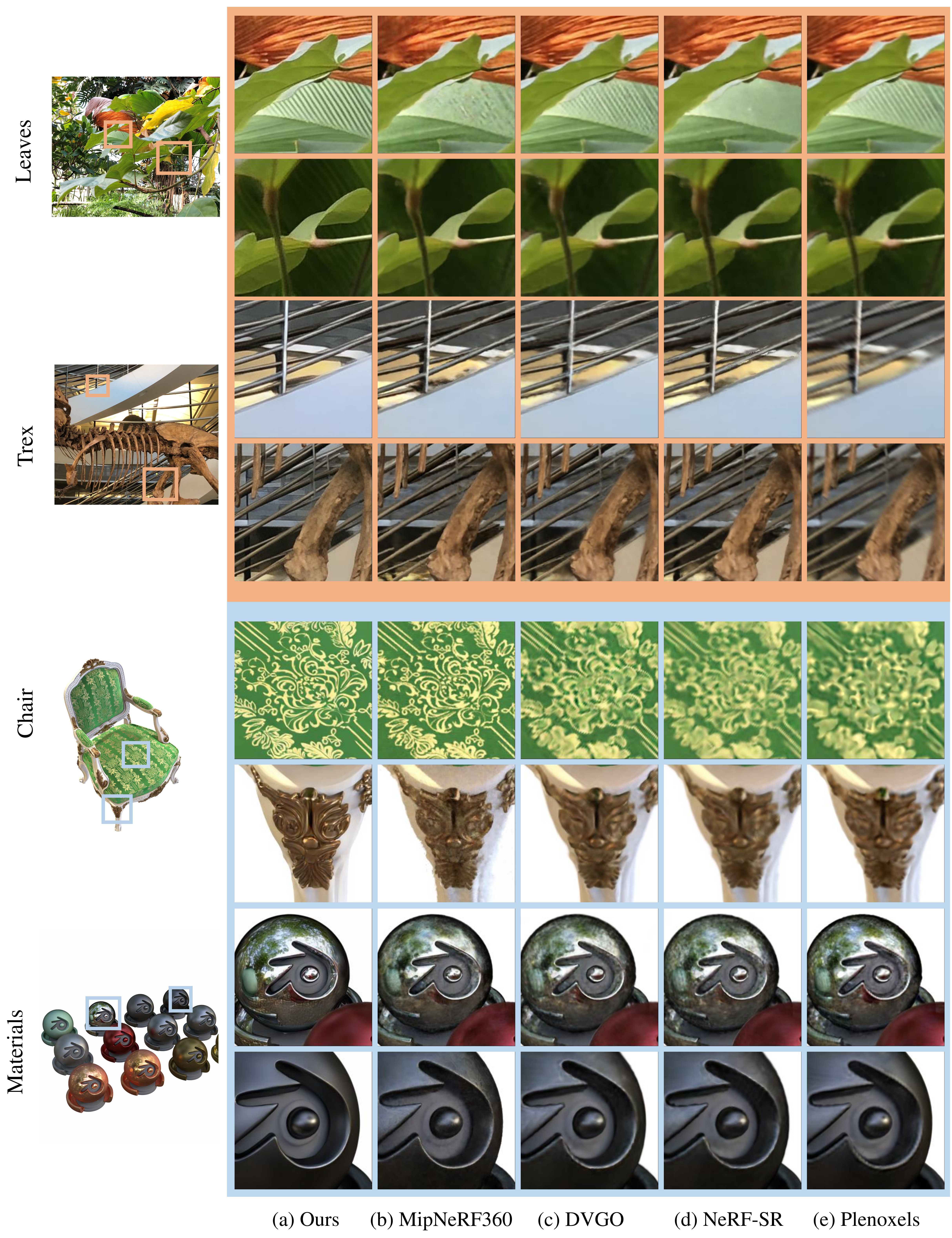}
   \caption{\textbf{Visual comparison with modern NeRF methods on example scenes from 4K-LLFF and 4K-Synthetic-NeRF.} Our method shows significant enhancement on preserving high-frequency details, either with complex geometry or high reflection surface, outperforming all the baseline methods obviously. The figure is better displayed on a high-resolution screen.}
   \label{fig:4kall}
\end{figure*}

\begin{figure}
  \centering
%   \fbox{\rule{0pt}{2in} \rule{0.9\linewidth}{0pt}}
   \includegraphics[width=0.7\linewidth]{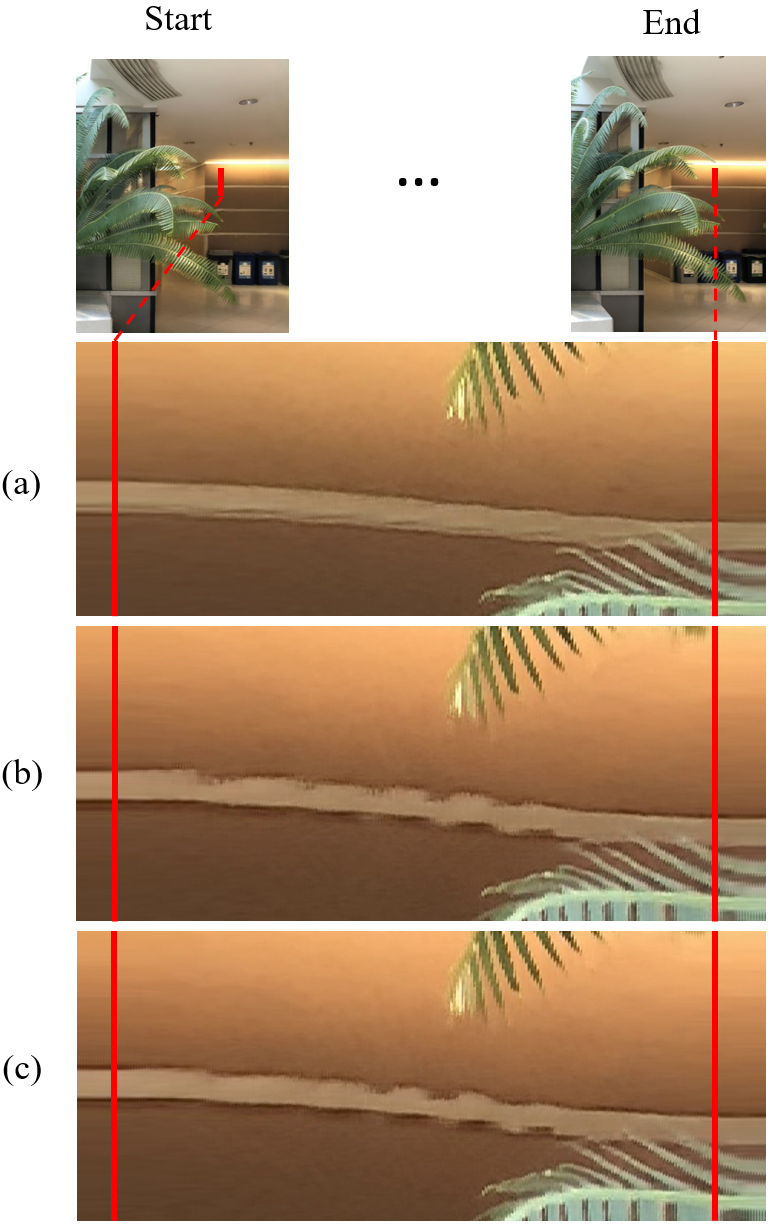}

   \caption{\textbf{View-consistency visualization}. From horizontal view interpolation videos, we extract a short vertical segment pixel at fixed location every frame and stack them horizontally to compare view consistency between (a) full training, (b) w/o depth modulation and (c) w/o joint training.}
   %(c) w/o joint training.}
   \label{fig:viewcs}
   \vspace{-1.0em}
\end{figure}

% \begin{figure}[t]
%   \centering
% %   \fbox{\rule{0pt}{2in} \rule{0.9\linewidth}{0pt}}
%   \includegraphics[width=1\linewidth]{figure/1kcp.png}

%     \caption{\textbf{Visual comparison on default 1K resolution LLFF dataset.} Our methods still achieve the best visual results compared to other methods. \llz{no more 1k resul, remeber to remove this figure after delete related text} }
%   \label{fig:1kcp}
%   \vspace{-0.5em}
% \end{figure}

\subsection{Evaluation Metrics}
 PSNR for evaluating distortion is used as the default metric in NeRF methods, while the metric is insensitive for the artifacts like over-smooth or blurry details, which has been well-analyzed in \cite{blau2018perception}. We hence evaluate the method with more metrics, including LPIPS \cite{zhang2018unreasonable} and NIQE \cite{mittal2012making} metrics for assessing perceptual effect, as well as another distortion-oriented metric SSIM \cite{SSIM}. LPIPS is calculated with AlexNet \cite{alexnet}.

\subsection{Datasets}
\textit{4K-LLFF.} The LLFF dataset \cite{BenMildenhall2019LocalLF} provides forward real-world scenes with training views at 4K ultra high resolution. 
%We use it to conduct experiments and ablation studies by default. 
It is composed of 8 forward-facing scenes and different scenes have different numbers of training views, between 20 and 60. Unlike using down-sampled images with 1K resolution in conventional NeRF methods, we use the full scale images  ($4032 \times 3024$)  for training and evaluation. 

% The original resolution is $4032 \times 3024$ 
% while existing NeRF-based methods use $4\times$ down-scaled images ($1008 \times 756$) for training and inference. In our experiments, we use the original 4K images as groundtruth for training and evaluation in the  experiments. 
%We use corresponding low-dimension ones in the ablation study of assessing the effect of framework for visual quality improvement at different resolutions (i.e., 2K and 1K). We follow other methods to use the camera poses estimated by COLMAP \cite{schoenberger2016sfm}.
%To the best of our knowledge, only the LLFF dataset has 4K images in the novel view synthesis task. This dataset is composed of 8 forward-facing real-world scenes captured by high-resolution equipment. Different scenes have different numbers of images, but they are all between 20 and 60. The native resolution of each set of scenes is 4032x3024. However, limited by training efficiency and quality, all previous methods used 4x downsampling (1008x756) images as ground truth. We followed other methods to estimate the approximate camera poses using COLMAP but used the original 4K images as the GT for all methods in the main experiment. Besides, we also use low-resolution GT in the ablation study to demonstrate the applicability of our method.

\textit{4K-Synthetic-NeRF.} The Synthetic-NeRF dataset \cite{NeRF} consists of the images rendered from 8 synthetic objects at the resolution of $800 \times 800$. Each scene contains $100$ training views and the other $200$ testing views. We re-render all the scene at $3200 \times 3200$ resolution based the original 3D models, forming the 4K version of the dataset.

\subsection{Comparisons}\label{ex:compare}
\textbf{Quantitative evaluation.} We first conduct the experiments to compare the method with modern NeRF methods, including Plenoxels \cite{yu2021plenoxels}, DVGO \cite{ChengSun2022DirectVG}, JaxNeRF \cite{jaxnerfgithub}, MipNeRF-360 \cite{mipnerf360} and NeRF-SR \cite{nerfsr} training and evaluating at 4K resolution. The results on 4K-LLFF and 4K-Synthetic-NeRF are respectively shown in Table~\ref{tab:main_comparison} and Table~\ref{tab:syn_comparison}. Our method (training with default loss setting) achieves obvious advantage in the perception metrics (i.e., LPIPS and NIQE) compared to all the baselines. The performance is comparable on the distortion metrics, slightly inferior to some baselines on the real-world scenes of LLFF. To better understand the method, we also provide the result by training a variant with $\mathcal{L}_1$ only in the decoder (i.e., without the adversarial and the perception losses) on LLFF, which achieves the best performance on the distortion metrics. Detailed analysis for the loss function can be found in the following ablation studies.  

Besides rendering quality metrics, we also provide inference time and training runtime memory as reference for a comprehensive evaluation. Our method achieves compelling performance on both metrics, allowing to render an 4K image within 600 ms. Our method achieves over $10\times$ faster inference with less than half training memory overhead compared to the direct counterpart DVGO.

\textbf{Qualitative comparison.} We provide the visual comparison in Fig.~\ref{fig:4kall}. Our method is capable of achieving high-fidelity photo-realistic rendering at such extremely high resolution scenes. The baseline methods show inferior ability on reconstructing subtle details at 4K scenes, incurring details lost or blur, e.g., leaf and chair texture. The visual quality of our method is obviously superior for preserving such complex and high-frequency details, even on the scenes with high reflection surfaces.

\newcommand{\tabincell}[2]{\begin{tabular}{@{}#1@{}}#2\end{tabular}}  

\begin{table*}[tbp]
\centering
\begin{tabular}{lcccccc}
\toprule
  & \multicolumn{2}{c}{Perception metrics}  &\multicolumn{2}{c}{Distortion metrics}   &  &   \\ 
\cmidrule(r){2-3} \cmidrule(r){4-5} 
\multirow{-2}{*}{Methods}     & LPIPS $\downarrow$    &NIQE$\downarrow$    &PSNR$\uparrow$  &SSIM$\uparrow$   & \multirow{-2}{*}{\tabincell{c}{Inference \\time (s)$\downarrow$}}   & \multirow{-2}{*}{\tabincell{c}{Runtime \\memory (GB)$\downarrow$}}   \\ 
\midrule
Plenoxels     & 0.48    & 8.86    &24.56  & 0.775 &1.88   &29.1   \\ 
DVGO          & 0.44    & 7.89    &25.13  & 0.779 &5.68   &58.6   \\ 
% Instant-NGP   & 0.44    & 7.82    &26.52  &  &25.075  & \textbf{5.5}   \\ 
JaxNeRF       & 0.42    & 7.03    & 25.37  & 0.773 &134.62 & 77.8   \\  
MipNeRF-360   & 0.37    & 6.31    &25.34  & 0.789 &51.38  & 78.1   \\ 
NeRF-SR       & 0.52    & 9.26    &24.15  & 0.754 &129.19 & 46.7   \\ 
\midrule
Ours-$\mathcal{L}_1$   &0.41            & 7.45          & \textbf{25.44}& \textbf{0.793} &0.58 &14.9\\
Ours      & \textbf{0.21}  &\textbf{4.75}  & 24.71  & 0.767 & \textbf{0.58} & \textbf{14.9}  \\  
\bottomrule
\end{tabular}
\caption{
    \textbf{Quantitative comparison with modern NeRF methods on 4K-LLFF dataset.} LPIPS is calculated with AlexNet. Our method ranks first on LPIPS and NIQE and achieve a comparable distortion performance. The variant of training only with $\ell_1$ loss can achieve better performance on distortion metrics. Our method also show benefits on inference speed and run-time memory overhead.
    }
\label{tab:main_comparison}
\end{table*}

% \begin{table}
% \centering
% \begin{tabular}{l|cccc}
    
% \toprule       
%     Method &  LPIPS$\downarrow$ &  NIQE$\downarrow$ &  PSNR$\uparrow$   &  SSIM$\uparrow$ \\  
%     \midrule
%     Plenoxels &  0.0974   & 8.94  &  29.45  & 0.937  \\
%     DVGO  &  0.0968   &  8.08    &   29.61  &  0.938 \\ 
%     % Instant-NGP & 0.0711   & 6.51   &  31.06  & 0.955 \\  
%     JaxNeRF   &  0.1020   &  6.98   &   29.98   & 0.928  \\ 
%     MipNeRF-360 &  0.0745  & 7.00  & 31.32   & 0.948 \\ 
%     NeRF-SR  & 0.1390  &   9.74    &   28.39   & 0.904  \\ 
%     \midrule
%     Ours  &  \textbf{0.0631}  &  7.50  & \textbf{30.71} & \textbf{0.952}  \\  
% \bottomrule
% \end{tabular}
% \caption{
%     { Comparison on 4K-Synthetic-NeRF dataset. \llz{remove NIQE add runtime memory}}
%     }
% \label{tab:syn_comparison}
% \vspace{-0.5em}
% \end{table}

\begin{table}
\setlength\tabcolsep{3pt}
\centering
% \small
\begin{tabular}{lcccc}
\toprule       
    &   &   & & Memory \\
    % Method &  LPIPS$\downarrow$ &  PSNR$\uparrow$   &  SSIM$\uparrow$ & Memory \\  

    \multirow{-2}{*}{Methods} & \multirow{-2}{*}{LPIPS$\downarrow$} & \multirow{-2}{*}{ PSNR$\uparrow$} &\multirow{-2}{*}{SSIM$\uparrow$} &(GB)$\downarrow$ \\
    \midrule
    Plenoxels &  0.097  &  29.45  & 0.937 & \textbf{10}  \\
    DVGO  &  0.097    &   29.61  &  0.938 & 48.4 \\ 
    % Instant-NGP & 0.0711   & 6.51   &  31.06  & 0.955 \\  
    JaxNeRF   &  0.102    &   29.98   & 0.928 & 77.7 \\ 
    MipNeRF-360 &  0.075   & \textbf{31.32}   & 0.948 & 77.2 \\ 
    NeRF-SR  & 0.139  &   28.39   & 0.904 & 46.7 \\ 
    \midrule
    Ours-$\mathcal{L}_1$  &  \textbf{0.063}  & 30.71 & \textbf{0.952} & 21.4 \\  
\bottomrule
\end{tabular}
\caption{
    {Quantitative comparison on 4K-Synthetic-NeRF.}
    %\llz{remove NIQE add runtime memory}
    }
\label{tab:syn_comparison}
\vspace{-0.5em}
\end{table}

\subsection{Ablation studies}
\textbf{Joint training.} \label{as: jointrain}
In order to better investigate the effect of joint training, we compare it to the setting of training the pair of encoder and decoder separately, i.e., fully train the encoder and then train the decoder without propagating gradient back to the encoder. We splice a clip of pixel strips at a fixed position in each frame of the rendered video and show the result in Fig.~\ref{fig:viewcs}. The margin of texture jitter is a strong indicator for judging consistency extent across view. Compared to joint training, there exists obvious texture jitter via separate training, showing that the rendering results with joint training are more view-consistent.

\begin{figure*}[t]
  \centering
   \includegraphics[width=1.0\linewidth]{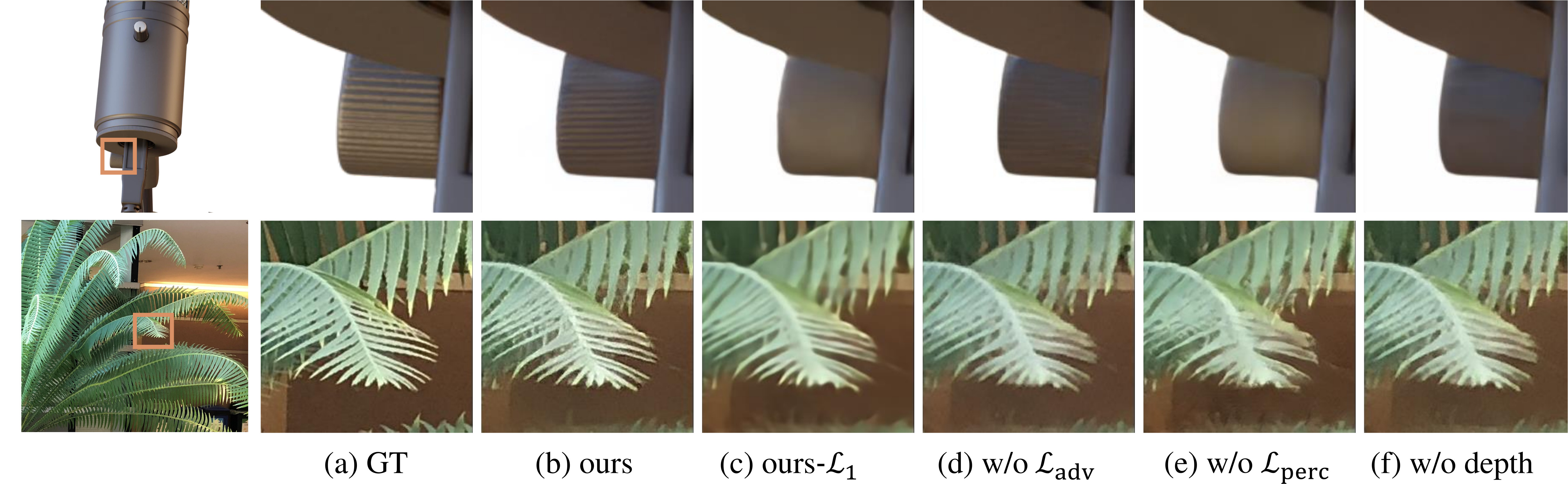}
    \vspace{-0.8em}
   \caption{Visual results of ablation studies on loss functions and depth modulation on the scenes of ``Mic" and ``Fern".}
   \label{fig:depth_ab}
\end{figure*}

\textbf{Depth modulation.} \label{as: depth} We integrate the explicit geometrical guidance into the decoder through the use of estimated depth, and validate its effect via the study without depth injection.
Modulation with depth can benefit rendering results with more view consistency compared to without depth (in Fig.~\ref{fig:viewcs}). It is helpful for improving rendering quality (as shown in Table~\ref{tab:ablation}) , especially for the scene details close to view plane, as shown in Fig.~\ref{fig:depth_ab}, which is more  consistent with human vision. 

\textbf{Loss function.} \label{as: loss} Using multiple losses would encourage the learning of discriminative patterns towards different aspects. As shown in Fig.~\ref{fig:depth_ab}, the regularization of the perception loss and the adversarial loss enables apparent visual quality improvement with richer and delicate textures (e.g., sharp leaf and screw thread) compared to using the distortion loss $\mathcal{L}_1$ only. Regularizing only with $\mathcal{L}_1$ may result in blurry and over-smooth artifacts on fine details although it can reach a higher value on the distortion metrics PSNR and SSIM. We also empirically found the adversarial loss shows better ability for recovering radiance compared to the perception loss.   
% \textbf{The capacity of decoder.} \label{as: decoder}
% In order to compare the effect of decoder capacity, we tested the recovery effect of four decoders with different configurations. From the experiment, the decoder capacity mainly affects the level of details but has no evident influence on the overall visual perception clarity of the image quality. The larger the decoder capacity, the more detail can be recovered. The tip of the leaf in Fig.~\ref{} becomes blurred as the model's size decreases and the texture of the microphone varies according to the model capacity. This difference in detail may not be apparent when the perspective moves.

% \begin{figure}[t]
%   \centering
% %   \fbox{\rule{0pt}{2in} \rule{0.9\linewidth}{0pt}}
%   \includegraphics[width=0.75\linewidth]{figure/depth_ab.png}
%     \vspace{-0.5em}
%   \caption{\textbf{Ablation of depth modulation.} The clarity of complex and subtle object details can be obviously improved after adding depth modulation.}
%   \label{fig:depth_ab}
% \end{figure}

\begin{figure}[t]
  \centering
  \includegraphics[width=0.7\linewidth]{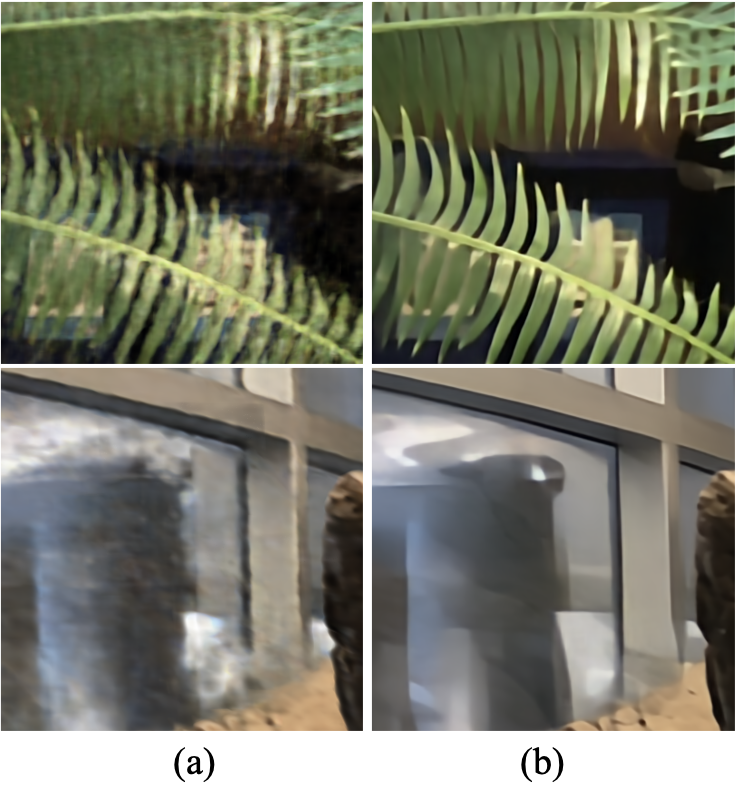}
  \vspace{-0.8em}
  \caption{Qualitative comparison of baseline TensoRF (a) and our method with TensoRF-based Encoder (b).}
  \label{fig:trf}
\end{figure}

\begin{table}
\centering
\begin{tabular}{lcccc}
    
\toprule       
    Method &  LPIPS$\downarrow$           &  NIQE$\downarrow$  &  PSNR$\uparrow$   &  SSIM$\uparrow$ \\  
    \midrule
    ours    & \textbf{0.162} &  \textbf{4.20}  &  23.49 & 0.771  \\
    ours-$\mathcal{L}_{\text{1}}$ & 0.353   &  6.37   &   \textbf{23.69}   &  \textbf{0.778}  \\ 
    % \midrule
        % \textsuperscript{$\llcorner$}Large & & & & &
    w/o depth  &  0.189     & 4.61     &   23.36   &  0.754 \\ 
    w/o $\mathcal{L}_{\text{adv}}$  &  0.205 & 6.89   &   23.39  &  0.759 \\  
    w/o $\mathcal{L}_{\text{perc}}$  &  0.241  &  4.51    & 23.31   & 0.764  \\ 
    
\bottomrule
\end{tabular}
\caption{
    {Ablation study of losses and depth on ``Fern" scene.}
    }
\label{tab:ablation}
\vspace{-0.5em}
\end{table}

\textbf{Encoder Backbone.} The base encoder can be instantiated with different NeRF-based architectures. In order to assess the generalization of our framework, we conduct the experiment by using TensoRF \cite{TensoRF} instead of DVGO as the encoder base. The qualitative and quantitative results are shown in Fig.~\ref{fig:trf} and Table.~\ref{tab:trf}. Clear improvements are achieved on both evaluation metrics and visual qualities, showing that our method can boost rendering quality on fine details and reduce blurry artifacts even on challenging transparent/translucent objects.

Other ablation studies, such as the impact of decoder capacity and patch size, are shown in supplementary materials.

\subsection{Limitation and Future Work}
Our method can recover high-frequency details well at ultra high-resolution scenes and show strong adaptability to reflection and translucency. However, we empirically found adding perception loss may happen to confine the recovery of highlight colors (e.g., slight color aberration on the button shown in Fig.\ref{fig:viewcs} top row). This may be alleviated by tuning the loss parameters in an elaborate manner or incorporating more meta information (e.g., ray direction, normal map and reflectance) into the decoder learning. The training time of our method is relatively long due to training the decoder with convolutional networks and the discriminator in the adversarial loss. As a future work we will consider taking advantage of pre-training models from image super-resolution tasks or extending the generalization of decoder (associated with fast fine-tuning) to reduce per-scene training cost.

\begin{table}[t]
\centering
\begin{tabular}{lcccc}
\toprule  
Scene                                           & Method  & LPIPS $\downarrow$                        & NIQE $\downarrow$                       & PSNR $\uparrow$                        \\ \midrule
                        & TensoRF & 0.464                        & 7.172                        & 23.33                        \\ %\cline{2-5} 
\multirow{-2}{*}{Fern}  & Ours    & {\color[HTML]{000000} 0.342} & {\color[HTML]{000000} 6.089} & {\color[HTML]{000000} 23.27} \\ \hline
                        & TensoRF & 0.452                        & 7.051                        & 26.19                        \\ %\cline{2-5} 
\multirow{-2}{*}{Horns} & Ours    & 0.387                        & 6.276                        & 26.72                        \\ 
\bottomrule
\end{tabular}
\caption{3D-Aware Encoder based on TensoRF.}
\label{tab:trf}
\end{table}

\begin{comment}
\textbf{GAN loss.} \label{as: gan}
GAN loss has been shown in multi-domains to increase detail and perceptual sharpness, which is vital for 4K images. We used patch-GAN loss\cite{isola2017image} to improve the details of images. This loss judges the difference between the output patches of VC-Decoder and GT. The training is complete when the discriminator cannot distinguish between the two patches. GAN loss increases the clarity of high-frequency details.
\end{comment}
\section{Conclusion}\label{conclusion}
In this paper, we explored the ability of NeRF methods on modelling fine details of 3D scenes and proposed a novel framework to boost its representational power on recovering subtle details at 4K ultra high resolutions. A pair of encoder-decoder modules are introduced to take better use of geometric properties for realizing impressive rendering quality on complex and high-frequency details, by virtue of local correlation captured from geometry-aware features. Patch-based sampling allows the training to integrate the supervision from perception-oriented regularization beyond pixel-level mechanism. We expect to investigate the effect of enhancing ray correlation, especially incorporated with the success of existing perception and generative methods, on pursing high-fidelity 3D scene modelling and manipulation as well as extending to dynamic scenes as future directions.

%%%%%%%%% REFERENCES
\bibliographystyle{ieee_fullname}
\bibliography{egbib}

\begin{thebibliography}{10}\itemsep=-1pt

\bibitem{mipnerf}
Jonathan~T. Barron, Ben Mildenhall, Matthew Tancik, Peter Hedman, Ricardo
  Martin-Brualla, and Pratul~P. Srinivasan.
\newblock Mip-nerf: A multiscale representation for anti-aliasing neural
  radiance fields.
\newblock {\em arXiv: Computer Vision and Pattern Recognition}, 2021.

\bibitem{mipnerf360}
Jonathan~T Barron, Ben Mildenhall, Dor Verbin, Pratul~P Srinivasan, and Peter
  Hedman.
\newblock Mip-nerf 360: Unbounded anti-aliased neural radiance fields.
\newblock In {\em Proceedings of the IEEE/CVF Conference on Computer Vision and
  Pattern Recognition}, pages 5470--5479, 2022.

\bibitem{blau2018perception}
Yochai Blau and Tomer Michaeli.
\newblock The perception-distortion tradeoff.
\newblock In {\em Proceedings of the IEEE conference on computer vision and
  pattern recognition}, pages 6228--6237, 2018.

\bibitem{MichaelBroxton2020ImmersiveLF}
Michael Broxton, John Flynn, Ryan Overbeck, Daniel Erickson, Peter Hedman,
  Matthew DuVall, Jason Dourgarian, Jay Busch, Matt Whalen, and Paul Debevec.
\newblock Immersive light field video with a layered mesh representation.
\newblock {\em ACM Transactions on Graphics}, 2020.

\bibitem{eg3d}
Eric~R. Chan, Connor~Z. Lin, Matthew~A. Chan, Koki Nagano, Boxiao Pan,
  Shalini~De Mello, Orazio Gallo, Leonidas Guibas, Jonathan Tremblay, Sameh
  Khamis, Tero Karras, and Gordon Wetzstein.
\newblock Efficient geometry-aware {3D} generative adversarial networks.
\newblock In {\em arXiv}, 2021.

\bibitem{TensoRF}
Anpei Chen, Zexiang Xu, Andreas Geiger, Jingyi Yu, and Hao Su.
\newblock Tensorf: Tensorial radiance fields.
\newblock In {\em In Proceedings of the European Conference on Computer
  Vision}, 2022.

\bibitem{AnpeiChen2021MVSNeRFFG}
Anpei Chen, Zexiang Xu, Fuqiang Zhao, Xiaoshuai Zhang, Fanbo Xiang, Jingyi Yu,
  and Hao Su.
\newblock Mvsnerf: Fast generalizable radiance field reconstruction from
  multi-view stereo.
\newblock In {\em International Conference on Computer Vision}, 2021.

\bibitem{dai2019second}
Tao Dai, Jianrui Cai, Yongbing Zhang, Shu-Tao Xia, and Lei Zhang.
\newblock Second-order attention network for single image super-resolution.
\newblock In {\em Proceedings of the IEEE/CVF conference on computer vision and
  pattern recognition}, pages 11065--11074, 2019.

\bibitem{jaxnerfgithub}
Boyang Deng, Jonathan~T. Barron, and Pratul~P. Srinivasan.
\newblock {JaxNeRF}: an efficient {JAX} implementation of {NeRF}.
\newblock
  \url{https://github.com/google-research/google-research/tree/master/jaxnerf},
  2020.

\bibitem{deng2022depth}
Kangle Deng, Andrew Liu, Jun-Yan Zhu, and Deva Ramanan.
\newblock Depth-supervised nerf: Fewer views and faster training for free.
\newblock In {\em Proceedings of the IEEE/CVF Conference on Computer Vision and
  Pattern Recognition}, pages 12882--12891, 2022.

\bibitem{dong2015image}
Chao Dong, Chen~Change Loy, Kaiming He, and Xiaoou Tang.
\newblock Image super-resolution using deep convolutional networks.
\newblock {\em IEEE transactions on pattern analysis and machine intelligence},
  38(2):295--307, 2015.

\bibitem{JohnFlynn2019DeepViewVS}
John Flynn, Michael Broxton, Paul Debevec, Matthew DuVall, Graham Fyffe, Ryan
  Overbeck, Noah Snavely, and Richard Tucker.
\newblock Deepview: View synthesis with learned gradient descent.
\newblock {\em Proceedings of the IEEE/CVF International Conference on Computer
  Vision}, 2019.

\bibitem{fastnerf}
Stephan~J Garbin, Marek Kowalski, Matthew Johnson, Jamie Shotton, and Julien
  Valentin.
\newblock Fastnerf: High-fidelity neural rendering at 200fps.
\newblock In {\em Proceedings of the IEEE/CVF International Conference on
  Computer Vision}, pages 14346--14355, 2021.

\bibitem{SNeRG}
Peter Hedman, Pratul~P. Srinivasan, Ben Mildenhall, Jonathan~T. Barron, and
  Paul~E. Debevec.
\newblock Baking neural radiance fields for real-time view synthesis.
\newblock In {\em Proceedings of the IEEE/CVF International Conference on
  Computer Vision}, 2021.

\bibitem{hu2021worldsheet}
Ronghang Hu, Nikhila Ravi, Alexander~C Berg, and Deepak Pathak.
\newblock Worldsheet: Wrapping the world in a 3d sheet for view synthesis from
  a single image.
\newblock In {\em Proceedings of the IEEE/CVF International Conference on
  Computer Vision}, pages 12528--12537, 2021.

\bibitem{hu2022efficientnerf}
Tao Hu, Shu Liu, Yilun Chen, Tiancheng Shen, and Jiaya Jia.
\newblock Efficientnerf efficient neural radiance fields.
\newblock In {\em Proceedings of the IEEE/CVF Conference on Computer Vision and
  Pattern Recognition}, pages 12902--12911, 2022.

\bibitem{ji2020real}
Xiaozhong Ji, Yun Cao, Ying Tai, Chengjie Wang, Jilin Li, and Feiyue Huang.
\newblock Real-world super-resolution via kernel estimation and noise
  injection.
\newblock In {\em proceedings of the IEEE/CVF conference on computer vision and
  pattern recognition workshops}, pages 466--467, 2020.

\bibitem{alexnet}
Alex Krizhevsky, Ilya Sutskever, and Geoffrey~E Hinton.
\newblock Imagenet classification with deep convolutional neural networks.
\newblock In {\em Advances in Neural Information Processing Systems}, 2012.

\bibitem{ledig2017photo}
Christian Ledig, Lucas Theis, Ferenc Husz{\'a}r, Jose Caballero, Andrew
  Cunningham, Alejandro Acosta, Andrew Aitken, Alykhan Tejani, Johannes Totz,
  Zehan Wang, et~al.
\newblock Photo-realistic single image super-resolution using a generative
  adversarial network.
\newblock In {\em Proceedings of the IEEE conference on computer vision and
  pattern recognition}, pages 4681--4690, 2017.

\bibitem{liang2021swinir}
Jingyun Liang, Jiezhang Cao, Guolei Sun, Kai Zhang, Luc Van~Gool, and Radu
  Timofte.
\newblock Swinir: Image restoration using swin transformer.
\newblock In {\em Proceedings of the IEEE/CVF International Conference on
  Computer Vision}, pages 1833--1844, 2021.

\bibitem{nsvf}
Lingjie Liu, Jiatao Gu, Kyaw~Zaw Lin, Tat-Seng Chua, and Christian Theobalt.
\newblock Neural sparse voxel fields.
\newblock In {\em Advances in Neural Information Processing Systems}, 2020.

\bibitem{OpticalModel}
Nelson~L. Max.
\newblock Optical models for direct volume rendering.
\newblock {\em {IEEE} Trans. Vis. Comput. Graph.}, 1995.

\bibitem{BenMildenhall2019LocalLF}
Ben Mildenhall, Pratul~P. Srinivasan, Rodrigo Ortiz-Cayon, Nima~Khademi
  Kalantari, Ravi Ramamoorthi, Ren Ng, and Abhishek Kar.
\newblock Local light field fusion: Practical view synthesis with prescriptive
  sampling guidelines.
\newblock {\em ACM Transactions on Graphics}, 2019.

\bibitem{NeRF}
Ben Mildenhall, Pratul~P. Srinivasan, Matthew Tancik, Jonathan~T. Barron, Ravi
  Ramamoorthi, and Ren Ng.
\newblock Nerf: Representing scenes as neural radiance fields for view
  synthesis.
\newblock In {\em In Proceedings of the European Conference on Computer
  Vision}, 2020.

\bibitem{mittal2012making}
Anish Mittal, Rajiv Soundararajan, and Alan~C Bovik.
\newblock Making a “completely blind” image quality analyzer.
\newblock {\em IEEE Signal processing letters}, 20(3):209--212, 2012.

\bibitem{MichaelNiemeyer2020GIRAFFERS}
Michael Niemeyer and Andreas Geiger.
\newblock Giraffe: Representing scenes as compositional generative neural
  feature fields.
\newblock In {\em Proceedings of the IEEE/CVF Conference on Computer Vision and
  Pattern Recognition}, 2021.

\bibitem{kilonerf}
Christian Reiser, Songyou Peng, Yiyi Liao, and Andreas Geiger.
\newblock Kilonerf: Speeding up neural radiance fields with thousands of tiny
  mlps.
\newblock In {\em Proceedings of the IEEE/CVF International Conference on
  Computer Vision}, pages 14335--14345, 2021.

\bibitem{roessle2022dense}
Barbara Roessle, Jonathan~T Barron, Ben Mildenhall, Pratul~P Srinivasan, and
  Matthias Nie{\ss}ner.
\newblock Dense depth priors for neural radiance fields from sparse input
  views.
\newblock In {\em Proceedings of the IEEE/CVF Conference on Computer Vision and
  Pattern Recognition}, pages 12892--12901, 2022.

\bibitem{shih20203d}
Meng-Li Shih, Shih-Yang Su, Johannes Kopf, and Jia-Bin Huang.
\newblock 3d photography using context-aware layered depth inpainting.
\newblock In {\em Proceedings of the IEEE/CVF Conference on Computer Vision and
  Pattern Recognition}, pages 8028--8038, 2020.

\bibitem{Simonyan15}
Karen Simonyan and Andrew Zisserman.
\newblock Very deep convolutional networks for large-scale image recognition.
\newblock In {\em International Conference on Learning Representations}, 2015.

\bibitem{DeepVoxelsLP}
Vincent Sitzmann, Justus Thies, Felix Heide, Matthias Nie{\ss}ner, Gordon
  Wetzstein, and Michael Zollh{\"o}fer.
\newblock Deepvoxels: Learning persistent 3d feature embeddings.
\newblock In {\em Computer Vision and Pattern Recognition}, 2019.

\bibitem{ChengSun2022DirectVG}
Cheng Sun, Min Sun, and Hwann-Tzong Chen.
\newblock Direct voxel grid optimization: Super-fast convergence for radiance
  fields reconstruction.
\newblock In {\em Proceedings of the IEEE/CVF Conference on Computer Vision and
  Pattern Recognition}, pages 5459--5469, 2022.

\bibitem{tucker2020single}
Richard Tucker and Noah Snavely.
\newblock Single-view view synthesis with multiplane images.
\newblock In {\em Proceedings of the IEEE/CVF Conference on Computer Vision and
  Pattern Recognition}, pages 551--560, 2020.

\bibitem{verbin2021refnerf}
Dor Verbin, Peter Hedman, Ben Mildenhall, Todd Zickler, Jonathan~T. Barron, and
  Pratul~P. Srinivasan.
\newblock {Ref-NeRF}: Structured view-dependent appearance for neural radiance
  fields.
\newblock {\em Proceedings of the IEEE/CVF International Conference on Computer
  Vision}, 2022.

\bibitem{nerfsr}
Chen Wang, Xian Wu, Yuan-Chen Guo, Song-Hai Zhang, Yu-Wing Tai, and Shi-Min Hu.
\newblock Nerf-sr: High quality neural radiance fields using supersampling.
\newblock In {\em Proceedings of the 30th ACM International Conference on
  Multimedia}, pages 6445--6454, 2022.

\bibitem{IBRNetLM}
Qianqian Wang, Zhicheng Wang, Kyle Genova, Pratul~P. Srinivasan, Howard Zhou,
  Jonathan~T. Barron, Ricardo Martin-Brualla, Noah Snavely, and Thomas
  Funkhouser.
\newblock Ibrnet: Learning multi-view image-based rendering.
\newblock In {\em Computer Vision and Pattern Recognition}, 2021.

\bibitem{wang2021real}
Xintao Wang, Liangbin Xie, Chao Dong, and Ying Shan.
\newblock Real-esrgan: Training real-world blind super-resolution with pure
  synthetic data.
\newblock In {\em Proceedings of the IEEE/CVF International Conference on
  Computer Vision}, pages 1905--1914, 2021.

\bibitem{basicsr}
Xintao Wang, Liangbin Xie, Ke Yu, Kelvin~C.K. Chan, Chen~Change Loy, and Chao
  Dong.
\newblock {BasicSR}: Open source image and video restoration toolbox.
\newblock \url{https://github.com/XPixelGroup/BasicSR}, 2022.

\bibitem{wang2018esrgan}
Xintao Wang, Ke Yu, Shixiang Wu, Jinjin Gu, Yihao Liu, Chao Dong, Yu Qiao, and
  Chen Change~Loy.
\newblock Esrgan: Enhanced super-resolution generative adversarial networks.
\newblock In {\em Proceedings of the European conference on computer vision
  (ECCV) workshops}, pages 0--0, 2018.

\bibitem{SSIM}
Zhou Wang, Alan~C. Bovik, Hamid~R. Sheikh, and Eero~P. Simoncelli.
\newblock Image quality assessment: from error visibility to structural
  similarity.
\newblock {\em IEEE transactions on image processing}, 2004.

\bibitem{wiles2020synsin}
Olivia Wiles, Georgia Gkioxari, Richard Szeliski, and Justin Johnson.
\newblock Synsin: End-to-end view synthesis from a single image.
\newblock In {\em Proceedings of the IEEE/CVF Conference on Computer Vision and
  Pattern Recognition}, pages 7467--7477, 2020.

\bibitem{Nex}
Suttisak Wizadwongsa, Pakkapon Phongthawee, Jiraphon Yenphraphai, and Supasorn
  Suwajanakorn.
\newblock Nex: Real-time view synthesis with neural basis expansion.
\newblock In {\em Proceedings of the IEEE/CVF International Conference on
  Computer Vision}, 2021.

\bibitem{YaoYao2018MVSNetDI}
Yao Yao, Zixin Luo, Shiwei Li, Tian Fang, and Long Quan.
\newblock Mvsnet: Depth inference for unstructured multi-view stereo.
\newblock In {\em ECCV}, 2018.

\bibitem{yu2021plenoxels}
Alex Yu, Sara Fridovich-Keil, Matthew Tancik, Qinhong Chen, Benjamin Recht, and
  Angjoo Kanazawa.
\newblock Plenoxels: Radiance fields without neural networks.
\newblock {\em arXiv preprint arXiv:2112.05131}, 2021.

\bibitem{yu2021plenoctrees_iccv}
Alex Yu, Ruilong Li, Matthew Tancik, Hao Li, Ren Ng, and Angjoo Kanazawa.
\newblock {PlenOctrees} for real-time rendering of neural radiance fields.
\newblock In {\em Proceedings of the IEEE/CVF International Conference on
  Computer Vision}, 2021.

\bibitem{yu2021pixelnerf}
Alex Yu, Vickie Ye, Matthew Tancik, and Angjoo Kanazawa.
\newblock pixelnerf: Neural radiance fields from one or few images.
\newblock In {\em Proceedings of the IEEE/CVF Conference on Computer Vision and
  Pattern Recognition}, pages 4578--4587, 2021.

\bibitem{zhang2021designing}
Kai Zhang, Jingyun Liang, Luc Van~Gool, and Radu Timofte.
\newblock Designing a practical degradation model for deep blind image
  super-resolution.
\newblock In {\em Proceedings of the IEEE/CVF International Conference on
  Computer Vision}, pages 4791--4800, 2021.

\bibitem{zhang2018unreasonable}
Richard Zhang, Phillip Isola, Alexei~A Efros, Eli Shechtman, and Oliver Wang.
\newblock The unreasonable effectiveness of deep features as a perceptual
  metric.
\newblock In {\em Proceedings of the IEEE conference on computer vision and
  pattern recognition}, pages 586--595, 2018.

\bibitem{zhang2018image}
Yulun Zhang, Kunpeng Li, Kai Li, Lichen Wang, Bineng Zhong, and Yun Fu.
\newblock Image super-resolution using very deep residual channel attention
  networks.
\newblock In {\em Proceedings of the European conference on computer vision
  (ECCV)}, pages 286--301, 2018.

\end{thebibliography}

\clearpage
\appendix

% --- PDF will be split by an editor (e.g. macOS preview), so need to restart from page 1
% \setcounter{page}{1}

% --- repeat the title (AT: haven't found a more elegant way to do this...)
\twocolumn[
\centering
\Large
\textbf{4K-NeRF: High Fidelity Neural Radiance Fields at Ultra High Resolutions} \\
\vspace{0.5em}Appendix \\
\vspace{1.0em}
] %< twocolumn
\appendix
\section{Details of Model Structure}

% \begin{figure}[t]
%   \centering
%   \includegraphics[width=1\linewidth]{figure/trf_cp.png}
%   \caption{Visual comparison based on TensoRF in 4K scenes. (a) TensoRF. (b) Our results.}
%   \label{fig:trf}
% \end{figure}

\begin{figure*}[t]
  \centering
  \includegraphics[width=1.0\linewidth]{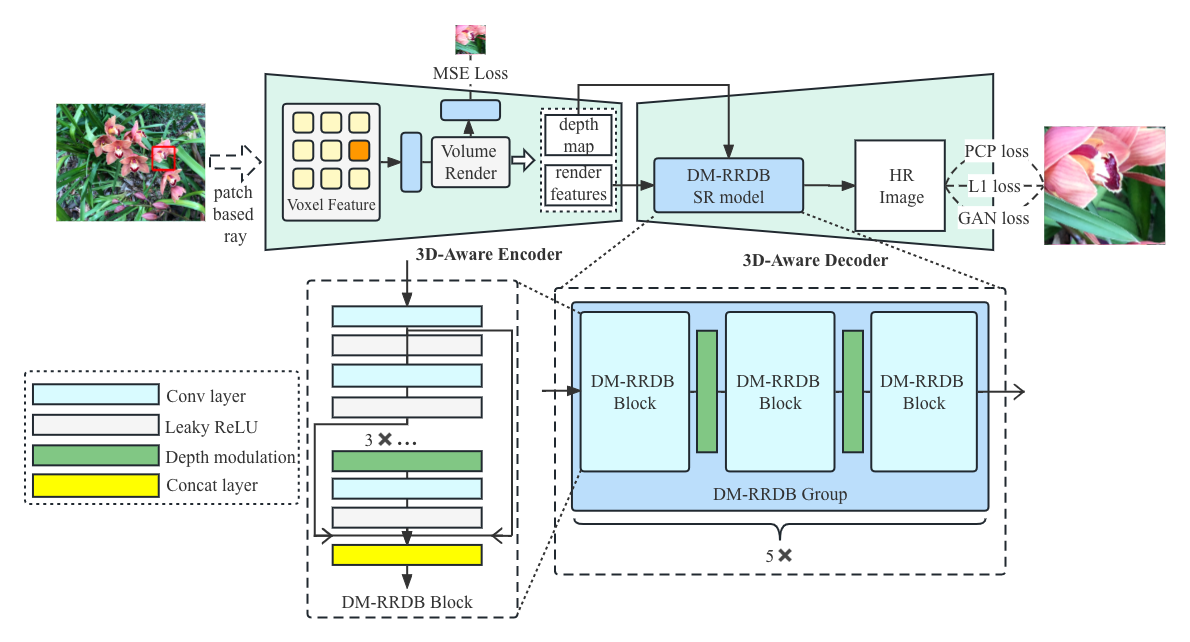}
  \caption{The scheme of 4K-NeRF in detail.}
  \label{fig:net}
\end{figure*}

% \begin{figure*}
%   \centering
%   \includegraphics[width=\linewidth]{figure_supp/pc_cp.pdf}
%   \caption{Comparison between different patch sizes. \llz{we need a better image for this ablation}}
%   \label{fig:pc_cp}
% \end{figure*}

% \begin{figure*}[t]
%   \centering
%   \begin{subfigure}{1.0\linewidth}
%     % \fbox{\rule{0pt}{2in} \rule{.9\linewidth}{0pt}}
%     \includegraphics[width=1.0\linewidth]{figure/loss_orchids.png}
%     \caption{Orchids scene in LLFF dataset.}
%     \label{fig:orchids}
%   \end{subfigure}
%   \hfill
%   \begin{subfigure}{1.0\linewidth}
%     % \fbox{\rule{0pt}{2in} \rule{.9\linewidth}{0pt}}
%     \includegraphics[width=1.0\linewidth]{figure/loss_room.png}
%     \caption{Room scene in LLFF dataset.}
%     \label{fig:room}
%   \end{subfigure}
%   \caption{Visual comparison of GAN loss and L1 loss on different scenes.}
%   \label{fig:loss_cp}
% \end{figure*}

We use the default configuration of DVGO \cite{ChengSun2022DirectVG} as the encoder setting in the experiments. Specifically, the size of voxels is $384 \times 384 \times 256$, and each voxel contains a density value representing geometry and a 12-dimensional color feature followed by a MLP. We extract ray features from the MLP with the channel dimension 64 following a dimensional reduction layer with the channel dimension 6. The encoder is trained on the resolution $1008 \times 756$.
%So our volume rendering operates in a 6-dimensional feature space rather than a 3-dimensional RGB space. To improve the convergence speed, we first train DVGO with low-resolution images $(1008 \times 756)$ to initialize the encoder.

The illustration of the 4K-NeRF structure is shown in Fig.~\ref{fig:net}. The decoder consists of 5 residual-in-residual dense modules (RRDB) \cite{wang2018esrgan,basicsr} with depth modulation (DM-) as well as one super-resolution head. Each module is comprised of three DM-RRDB blocks interleaved with depth modulation units. We also insert a depth modulation unit for each DM-RRDB block. More detailed configuration can refer to the network configuration provided in the source code. Resolution increase performs in the super-resolution head by stacking two convolutional layers interleaved with  $2\times$ bi-linear upsamling operation.

\begin{table*}[t]
\centering
\begin{tabular}{lcccccc}
    
\toprule       
    Patch Size &  LPIPS$\downarrow$           &  NIQE$\downarrow$  &  PSNR$\uparrow$   &  SSIM$\uparrow$ &  Training time(min)$\downarrow$  &  Runtime Memory(GB)$\downarrow$ \\  
    \midrule
    32 & 0.24 & 5.93 & 24.80 & 0.767  & 240 & 13.8 \\
    64 & 0.21 & 4.75 & 24.71 & 0.767  & 300 & 14.9 \\ 
    % \midrule
        % \textsuperscript{$\llcorner$}Large & & & & &
    128  & 0.18 & 5.33 & 24.85 & 0.760  & 600 & 17.6 \\ 
    256  & 0.19 & 5.27 & 24.70 & 0.757  & 780 & 29.1 \\  
    
\bottomrule
\end{tabular}
\caption{
    {Ablation study of different patch size on 4K-LLFF.}
    }
\label{tab:patch_size}
\vspace{-0.5em}
\end{table*}

\begin{table}[t]
\setlength\tabcolsep{1.8pt}
\centering
\begin{tabular}{llccc}
\toprule  
    &   &   &   &  Memory \\
    \multirow{-2}{*}{Dataset} &  \multirow{-2}{*}{Method} & \multirow{-2}{*}{LPIPS$\downarrow$} & \multirow{-2}{*}{ PSNR$\uparrow$}  &(GB)$\downarrow$ \\
\midrule
  & DVGO            & 0.44          &25.13           & 58.6  \\
  & DVGO$_{large}$  & 0.39          & 25.53         & 72.6   \\ %\cline{2-5} 
\multirow{-2}{*}{LLFF}  & Ours-$\mathcal{L}_1$     & 0.41  &  25.44  & 14.9  \\ 

    & Ours            & 0.21         & 24.71          & 14.9   \\ \hline
   & DVGO            & 0.10         & 29.61          & 48.4  \\
   & DVGO$_{large}$  & 0.07         & 31.42          & 77.2   \\ %\cline{2-5} 
\multirow{-2}{*}{4K-SYN}   & Ours-$\mathcal{L}_1$       &   0.06         & 30.71           & 21.4  \\ 
  & Ours            & 0.03         & 29.12           & 21.4   \\ 
    
\bottomrule
\end{tabular}
\caption{Quantitative comparison on 4K-LLFF and 4K-Synthetic-NeRF datasets among DVGO, DVGO$_{large}$ and ours.}
\label{tab:dv}
\vspace{-0.5em}
\end{table}

\begin{table*}
\centering
\begin{tabular}{l|ccccccc}
    
\toprule       
    Decoder setting &  LPIPS$\downarrow$           &  NIQE$\downarrow$  &  PSNR$\uparrow$   &  SSIM$\uparrow$ &  Training time$\downarrow$ &  Inference time$\downarrow$  &  Parameters \\  
    \midrule
    Large (5B64C) & 0.207 &  4.75  &  24.71 & 0.767  & 300 min &  0.58 s & 4.0 MB \\
    Medium (3B64C) & 0.216   &  5.12   &  24.47   &  0.759 & 255 min & 0.47 s  & 2.4 MB \\ 
    % \midrule
        % \textsuperscript{$\llcorner$}Large & & & & &
    Small (1B64C)  &  0.223    &  5.20    &  24.30  & 0.761 & 154 min & 0.31 s & 0.9 MB \\ 
    % 1B32C  &  0.240 &  5.95   &  24.62  & 0.760  &  & 240.179K \\  
    
\bottomrule
\end{tabular}
\caption{
    {Ablation study of different decoder size in 4K-LLFF datasets.}
    }
\label{tab:decoder_size}
\vspace{-0.5em}
\end{table*}

\section{More Descriptions on Evaluation Metrics}
Existing NeRF methods are typically supervised by pixel-level MSE loss and estimated by its direct counterpart PSNR metric. 
However, only using pixel-level loss is intractable to estimate problems like over-smooth details and blurry visual artifacts. These issues have been well analyzed and explained in detail in the papers \cite{blau2018perception,zhang2018unreasonable}, revealing the relation between perceptual quality and the degree of distortion. Distortion-oriented metrics (such as PSNR) can be treated as a visual lower bound, ensuring that semantic content in the image is consistent when reaching a certain level. The perceptual effects towards human vision, such as texture details and sharpness, can be measured by virtue of perception-oriented metrics, e.g., LPIPS. 
% Fig.~\ref{} shows some images with different visual qualities and their corresponding scores on PSNR and LPIPS, illuminating that PSNR may be inconsistent with visual quality estimated by human eyes. 
PSNR may be inconsistent with visual quality estimated by human eyes. This phenomenon is often more pronounced in ultra-high-resolution videos. Therefore, to quantify and compare the results more reasonably, we use LPIPS and NIQE as evaluation metrics besides PSNR. LPIPS and PSNR are calculated based on test ground-truth views (whose number is limited). As NIQE is a GT-free metric, we calculate across frames of rendered videos given camera trace to better assess cross-view quality.

%Because novel view synthesis methods such as NeRF need to obtain the observation effect from many different perspectives. However, LPIPS and PSNR can only be calculated under very few views with test ground truth. In order to better measure the quality of rendered videos from different views, we use the NIQE indicator, a GT-free metric. Therefore, we calculate each frame of the rendered video and use the final average score as an indicator of the video.

\section{More Ablation Studies} 
\textbf{The impact of Patch Size.}
We trained the model with the patches of four sizes and the qualitative results are shown in Table~\ref{tab:patch_size}. The method can achieve a comparable rendering quality across different patch sizes except using a relatively small patch size ($32 \times 32$). It may be less effective for capturing ray correlations from a restricted neighbouring context, resulting in inferior performance on perception metrics. On the other hand, using a larger patch size requires longer training to convergence as well as memory cost. Therefore we recommend choosing a moderate patch size (between 64 and 128), and used 64 by default in the experiments. 

\textbf{The impact of decoder capacity.} We conduct the ablation study on decoder capacity with the following three levels, ``small", ``medium" and ``large". The comparison results are shown in Table~\ref{tab:decoder_size}. Training a larger decoder costs longer while it shows better ability on improving visual quality. As the core motivation of the work is pursuing high-fidelity rendering, we use the large setting by default while it can achieve a trade-off between rendering quality and other performance metrics (e.g., training cost) by adjusting the decoder capacity.  

\begin{figure*}[p]
\begin{minipage}{\linewidth}
  \includegraphics[width=\linewidth]{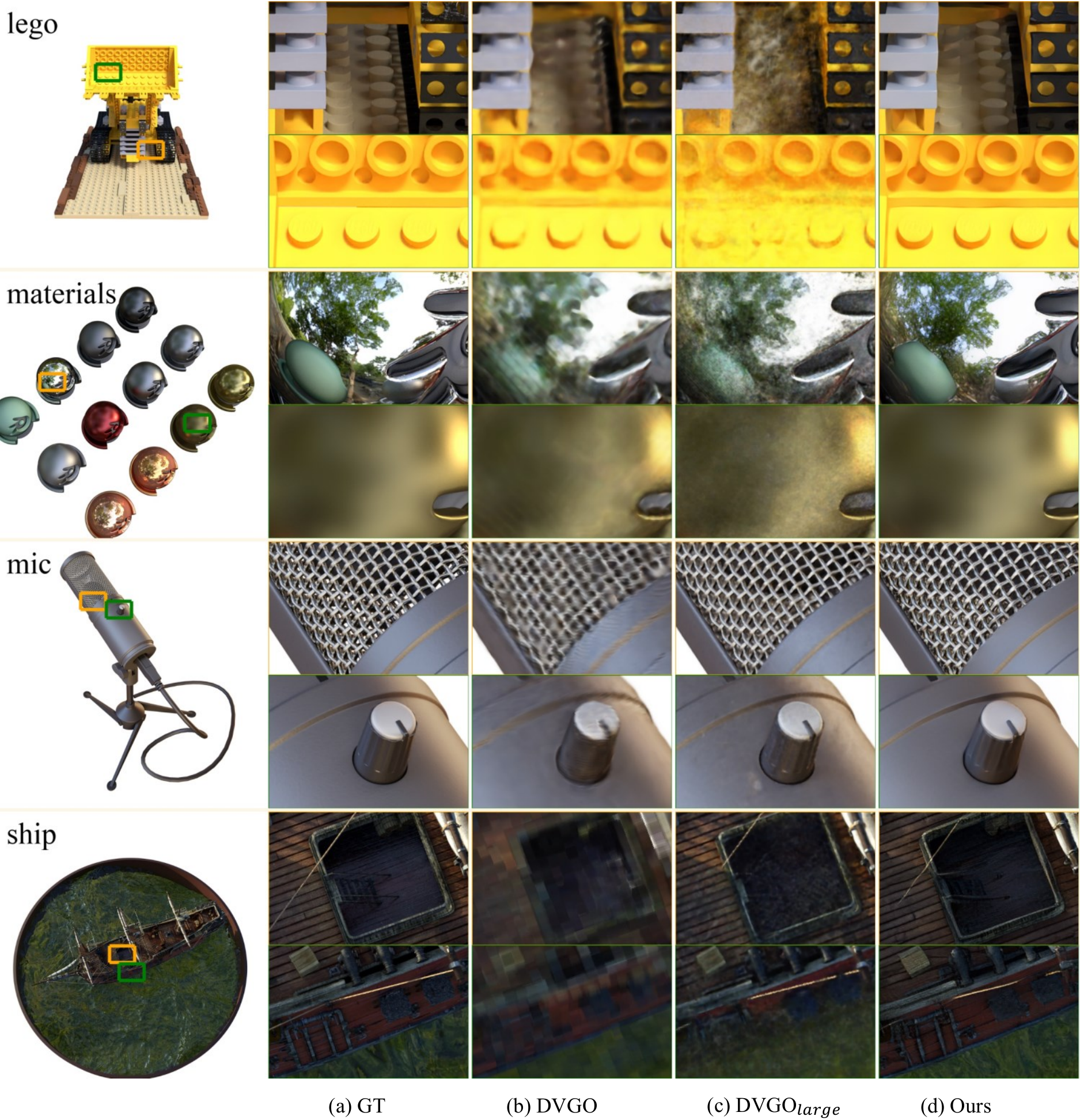}
  \caption{\textbf{Visual comparison with all kinds of baseline methods on each scenes from 4K-Synthetic-NeRF.} Our method shows significant enhancement on preserving high-frequency details, either with complex geometry or high reflection surface, outperforming DVGO and its variant obviously. }
  \label{fig:syn}
\end{minipage}
\end{figure*}

\textbf{Directly Scaling Up baseline.} 
We further investigate the ability of expanding traditional NeRF models to a larger capacity, and compare it with our method. We scale up the direct counterpart DVOG, named as  DVGO$_{large}$, by significantly increasing its model capacity up to running-time memory limit, i.e., increasing the number of MLP channels from 64 to 128, doubling the number of training epochs and expanding the dimension of voxel grids from $384 \times 384 \times 256$ to $1200 \times 1000 \times 256$ on the 4K-LLFF dataset and from $160^{3}$ to $640^{3}$ on the 4K-Synthetic-NeRF dataset. 
%Under this setting, the peak GPU memory consumption of DVGO$-{large}$ training almost exhausts a single 80G A100 card. 
The qualitative and visual comparison among DVGO, DVGO$_{large}$ and ours are shown in Table.~\ref{tab:dv}, Fig.~\ref{fig:syn}. Our method shows obvious improvement compared to DVGO$_\text{large}$ on visually detail recovery as well as the value of perception metric. We also found that compared to the standard setting of DVGO, the large variant sometimes exist more significant artifacts, e.g., cluttered textures and lack of leaves.  In contrast, our method can achieve consistent enhancement on visual quality, especially for high-frequency details.

\section{Detailed Results}
We present detailed results for each scene on the 4K-LLFF and 4K-Synthetic-NeRF datasets in  Tables \ref{tab:llff}. In addition, we provide rendered videos on four representative scenes (``Fern", ``Horns", ``Drums" and ``Mic") for better illustrating the superiority of our method on visual quality of 4K scenes, which we recommend to watch on the 4K ultra-high-resolution display.

\begin{table*}[h!]
\centering
\begin{tabular}{l|ccccccc}
\toprule
% \multicolumn{1}{c}{}     & &    & \multicolumn{2}{c}{Perception metrics}        & Distortion index&&\\ \cline{4-6}
% \multicolumn{1}{c}{\multirow{-2}{*}{Scene}} & \multirow{-2}{*}{Method}      & \multirow{-2}{*}{Setting}        & LPIPS\downarrow& NIQE\downarrow& PSNR\uparrow & \multirow{-2}{*}{\begin{tabular}[c]{@{}c@{}}Inference time\\ (s)\downarrow\end{tabular}} & \multirow{-2}{*}{\begin{tabular}[c]{@{}c@{}}Cache memory\\ (GB)\downarrow\end{tabular}} \\ 
\multirow{2}{*}{Scene} & \multirow{2}{*}{Method} & \multicolumn{2}{c}{Perception metrics}& \multicolumn{2}{c}{Distortion metrics}   & Inference time      & Cache memory\\ \cline{3-6} 
 &   &  LPIPS $\downarrow$ & NIQE $\downarrow$ & PSNR $\uparrow$ & SSIM $\uparrow$ & (s) $\downarrow$ & (GB) $\downarrow$ \\\midrule
 
    %  & & standard & 0.456     & 7.721    & 23.842    & 2.3    & 32.8\\ 
     & Plenoxels      & 0.456     & 7.721    & 23.842  & 0.772  & 2.3    & 32.8  \\ 
     
     & DVGO    & 0.424     & 6.910    & 23.741  & 0.771 & 6.2  & 20.1\\
    %  & DVGO-${large}$   & 0.346    & 5.543    & 23.684  & 0.772  & 12.6   & 68.3 \\
    %  & \multirow{-2}{*}{DVGO}        & large   & 0.346    & 5.543    & 23.684    & 12.6   & 68.3\\ \cline{2-8} 
     
     & JaxNeRF     & 0.399    & 5.623    & 23.470  & 0.758  & 134.7  & 77.8\\  
    %  & \multirow{-2}{*}{JaxNeRF}     & large        & 0.354    & 5.312    & 22.689    & 279.3  & 77.8\\ \cline{2-8} 
     
     & MipNeRF-360   & 0.348    & 5.229    & 23.867  & 0.786  & 51.3   & 78.1\\  
    %  & \multirow{-2}{*}{MipNeRF-360} & large        & 0.321    & 4.986    & 23.900    & 105.2  & 78.1\\ \cline{2-8} 
     
     & NeRF-SR     & 0.516    & 7.362    & 22.893  & 0.735  & 129.6  & 46.7\\ \cline{2-8}
     
     & Ours & 0.190 & 4.201 & 23.494  & 0.771 & 0.3 & 11.8\\  
\multirow{-7}{*}{Fern}  
    & Ours-$\mathcal{L}_1$  & 0.353    & 6.377    & 23.691  & 0.778  & 0.3  & 11.8\\ \bottomrule

\end{tabular}
\end{table*}

\begin{table*}[!h]
\centering
\begin{tabular}{l|ccccccc}
\toprule
\multirow{2}{*}{Scene} & \multirow{2}{*}{Method} & \multicolumn{2}{c}{Perception metrics}& \multicolumn{2}{c}{Distortion metrics}   & Inference time      & Cache memory\\ \cline{3-6} 
 &   &  LPIPS $\downarrow$ & NIQE $\downarrow$ & PSNR $\uparrow$ & SSIM $\uparrow$ & (s) $\downarrow$ & (GB) $\downarrow$ \\\midrule
 
     & Plenoxels  & 0.516    & 10.42   & 26.103  & 0.811  & 2.5    & 29.3\\   
    %  & \multirow{-2}{*}{Plenoxels}   & large        & 0.518    & 10.48   & 26.133    & 4.2    & 77.2\\ \cline{2-8} 
     & DVGO & 0.500    & 9.964    & 26.857   & 0.812 & 5.6  & 26.5\\  
    %  & DVGO-${large}$   & 0.456    & 8.931     & 27.368   & 0.818 & 10.7   & 78.4\\ 
     & JaxNeRF   & 0.489    & 9.308    & 26.783  & 0.806  & 134.7  & 77.8\\   
    %  & \multirow{-2}{*}{JaxNeRF}     & large        & 0.458    & 8.890    & 27.265    & 279.3  & 77.8\\ \cline{2-8} 
     & MipNeRF-360  & 0.437    & 7.824    & 27.119  &  0.812 & 51.3   & 78.1\\  
    %  & \multirow{-2}{*}{MipNeRF-360} & large        & 0.402    & 7.287    & 27.280    & 105.2  & 78.1\\ \cline{2-8} 
     & NeRF-SR    & 0.556    & 11.07   & 25.578  & 0.784  & 129.6  & 46.7\\ \cline{2-8} 
     & Ours  & 0.235 & 5.525 & 26.454 & 0.792 & 0.27 & 14.2\\  
\multirow{-7}{*}{Flower}
    & Ours-$\mathcal{L}_1$ & 0.493     & 9.514    & 26.865  & 0.820  & 0.27 & 14.2\\ \bottomrule
\end{tabular}
\end{table*}

\begin{table*}[!h]
\centering
\begin{tabular}{l|ccccccc}
\toprule
\multirow{2}{*}{Scene} & \multirow{2}{*}{Method} & \multicolumn{2}{c}{Perception metrics}& \multicolumn{2}{c}{Distortion metrics}   & Inference time      & Cache memory\\ \cline{3-6} 
 &   &  LPIPS $\downarrow$ & NIQE $\downarrow$ & PSNR $\uparrow$ & SSIM $\uparrow$ & (s) $\downarrow$ & (GB) $\downarrow$ \\\midrule
     & Plenoxels  & 0.491    & 9.919    & 28.852   & 0.860  & 2.4    & 30.1\\  
    %  & \multirow{-2}{*}{Plenoxels}   & large        & 0.488    & 9.937    & 28.854    & 4.1  & 76.1\\ \cline{2-8} 
     & DVGO  & 0.397    & 8.766    & 29.438   &  0.864  & 5.3  & 30.7\\  
    %  &  DVGO-${large}$        & 0.353    & 8.055    & 30.029   &  0.873 & 9.5    & 71.9\\ 
     & JaxNeRF    & 0.336    & 7.737    & 30.210  &  0.869  & 134.7  & 77.8\\  
    %  & \multirow{-2}{*}{JaxNeRF}     & large        & 0.334    & 7.490    & 29.882    & 279.3  & 77.8\\ \cline{2-8} 
     & MipNeRF-360    & 0.314    & 7.472    & 30.169  & 0.873   & 51.3   & 78.1\\  
    %  & \multirow{-2}{*}{MipNeRF-360} & large        & 0.294    & 6.806    & 30.231    & 105.2  & 78.1\\ \cline{2-8} 
     & NeRF-SR    & 0.517    & 9.637    & 28.719   & 0.859  & 129.6  & 46.7\\ \cline{2-8} 
     & Ours &  0.197 & 4.857 & 28.120 & 0.846 & 0.25 & 15.3\\  
\multirow{-7}{*}{Fortress}        
& Ours-$\mathcal{L}_1$   & 0.404    & 8.320    & 29.853   & 0.876  & 0.25 & 15.3\\ \bottomrule
\end{tabular}
\end{table*}

\begin{table*}[!p]
\centering
\begin{tabular}{l|ccccccc}
\toprule
\multirow{2}{*}{Scene} & \multirow{2}{*}{Method} & \multicolumn{2}{c}{Perception metrics}& \multicolumn{2}{c}{Distortion metrics}   & Inference time      & Cache memory\\ \cline{3-6} 
 &   &  LPIPS $\downarrow$ & NIQE $\downarrow$ & PSNR $\uparrow$ & SSIM $\uparrow$ & (s) $\downarrow$ & (GB) $\downarrow$ \\\midrule
     & Plenoxels & 0.510    & 8.298    & 24.743  &  0.756  & 2.3    & 31.0  \\  
    %  & \multirow{-2}{*}{Plenoxels}   & large        & 0.511    & 8.235    & 24.763    & 4.4  & 78.4\\ \cline{2-8} 
     & DVGO & 0.462    & 7.053    & 25.632   &  0.760  & 5.4  & 40.8\\  
    %  & DVGO-${large}$        & 0.394    & 6.340    & 26.367  &  0.784  & 9.6    & 72.0  \\ 
     & JaxNeRF     & 0.430    & 5.945    & 26.127  &  0.770  & 134.7  & 77.8\\  
    %  & \multirow{-2}{*}{JaxNeRF}     & large        & 0.402     & 5.853    & 26.760    & 279.3  & 77.8\\ \cline{2-8} 
     & MipNeRF-360   & 0.371    & 5.172    & 26.220  &  0.790  & 51.3   & 78.1\\  
    %  & \multirow{-2}{*}{MipNeRF-360} & large        & 0.344    & 4.952    & 26.224    & 105.2  & 78.1\\ \cline{2-8} 
     & NeRF-SR     & 0.553    & 9.758   & 23.694  &  0.743  & 129.6  & 46.7\\ \cline{2-8}
     &  Ours & 0.191 & 4.439 & 25.066 & 0.742 & 0.29 & 18.8\\  
\multirow{-7}{*}{Horns} 
& Ours-$\mathcal{L}_1$   & 0.399    & 6.241    & 26.336   & 0.794  & 0.29 & 18.8\\ \bottomrule
\end{tabular}
\end{table*}

\begin{table*}[!p]
\centering
\begin{tabular}{l|ccccccc}
\toprule
\multirow{2}{*}{Scene} & \multirow{2}{*}{Method} & \multicolumn{2}{c}{Perception metrics}& \multicolumn{2}{c}{Distortion metrics}   & Inference time      & Cache memory\\ \cline{3-6} 
 &   &  LPIPS $\downarrow$ & NIQE $\downarrow$ & PSNR $\uparrow$ & SSIM $\uparrow$ & (s) $\downarrow$ & (GB) $\downarrow$ \\\midrule
     & Plenoxels  & 0.520    & 7.749    & 20.028   & 0.661  & 1.0    & 23.3\\ 
    %  & \multirow{-2}{*}{Plenoxels}   & large        & 0.518    & 7.712   & 20.030    & 1.5  & 59.1\\ \cline{2-8} 
     & DVGO & 0.511    & 7.388   & 20.220  &  0.656  & 5.7  & 22.6\\  
    %  & DVGO-${large}$    & 0.447    & 6.568   & 20.211  &  0.663   & 10.1   & 71.0  \\ 
     & JaxNeRF    & 0.536    & 6.942   & 19.781  &  0.617  & 134.7  & 77.8\\  
    %  & \multirow{-2}{*}{JaxNeRF}     & large        & 0.453    & 6.294   & 20.148    & 279.3  & 77.8\\ \cline{2-8} 
     & MipNeRF-360   & 0.427    & 6.078   & 19.835  &  0.660  & 51.3   & 78.1\\  
    %  & \multirow{-2}{*}{MipNeRF-360} & large        & 0.379    & 5.593   & 19.970     & 105.2  & 78.1\\ \cline{2-8} 
     & NeRF-SR      & 0.559    & 8.167   & 19.033  &  0.604  & 129.6  & 46.7\\ \cline{2-8} 
     & Ours & 0.227 & 4.367 & 19.781 & 0.648 & 0.25 & 13.4\\  
\multirow{-7}{*}{Leaves}
&  Ours-$\mathcal{L}_1$     & 0.461    & 7.075   & 19.819  & 0.665   & 0.25 & 13.4\\ \bottomrule
\end{tabular}
\end{table*}

\begin{table*}[!h]
\centering
\begin{tabular}{l|ccccccc}
\toprule
\multirow{2}{*}{Scene} & \multirow{2}{*}{Method} & \multicolumn{2}{c}{Perception metrics}& \multicolumn{2}{c}{Distortion metrics}   & Inference time      & Cache memory\\ \cline{3-6} 
 &   &  LPIPS $\downarrow$ & NIQE $\downarrow$ & PSNR $\uparrow$ & SSIM $\uparrow$ & (s) $\downarrow$ & (GB) $\downarrow$ \\\midrule
     & Plenoxels  & 0.575    & 9.150    & 19.874   & 0.670 & 2.1    & 35.3\\  
    %  & \multirow{-2}{*}{Plenoxels}   & large        & 0.572    & 9.036    & 19.870     & 3.7  & 69.4\\ \cline{2-8} 
     & DVGO  & 0.539    & 8.112    & 20.098  & 0.670 & 6.1  & 22.5\\  
    %  & DVGO-${large}$    & 0.491    & 6.924    & 19.970  & 0.664 & 14.3   & 73.7\\ 
     & JaxNeRF   & 0.549    & 7.872    & 19.649  & 0.643  & 134.7  & 77.8\\  
    %  & \multirow{-2}{*}{JaxNeRF}     & large        & 0.498    & 7.147    & 19.383   & 279.3  & 77.8\\ \cline{2-8} 
     & MipNeRF-360     & 0.482    & 6.880    & 19.511  & 0.662 & 51.3   & 78.1\\  
    %  & \multirow{-2}{*}{MipNeRF-360} & large        & 0.432    & 6.234     & 19.672   & 105.2  & 78.1\\ \cline{2-8} 
     & NeRF-SR    & 0.594    & 8.973    & 19.432  & 0.637 & 129.6  & 46.7\\ \cline{2-8} 
     & Ours & 0.236 & 5.203 & 20.005 & 0.649 & 0.32 & 12.5\\  
\multirow{-7}{*}{Orchids}
& Ours-$\mathcal{L}_1$       & 0.523    & 7.649    & 19.557  & 0.670  & 0.32 & 12.5\\ \bottomrule
\end{tabular}
\end{table*}

\begin{table*}[!h]
\centering
\begin{tabular}{l|ccccccc}
\toprule
\multirow{2}{*}{Scene} & \multirow{2}{*}{Method} & \multicolumn{2}{c}{Perception metrics}& \multicolumn{2}{c}{Distortion metrics}   & Inference time      & Cache memory\\ \cline{3-6} 
 &   &  LPIPS $\downarrow$ & NIQE $\downarrow$ & PSNR $\uparrow$ & SSIM $\uparrow$ & (s) $\downarrow$ & (GB) $\downarrow$ \\\midrule
     & Plenoxels  & 0.392    & 9.038     & 28.133   & 0.884  & 2.7    & 20.5\\  
    %  & \multirow{-2}{*}{Plenoxels}   & large        & 0.390    & 9.038    & 28.133    & 5.2  & 75.3\\ \cline{2-8} 
     & DVGO  & 0.357    & 7.979    & 29.554  &  0.893  & 5.4  & 30.3\\  
    %  & DVGO-${large}$     & 0.323    & 7.486    & 30.834  &  0.905  & 9.4    & 71.4\\ 
     & JaxNeRF   & 0.317    & 6.744    & 31.114   & 0.908  & 134.7  & 77.8\\  
    %  & \multirow{-2}{*}{JaxNeRF}     & large        & 0.309    & 7.360    & 31.733    & 279.3  & 77.8\\ \cline{2-8} 
     & MipNeRF-360    & 0.309    & 6.614    & 30.687  &  0.908  & 51.3   & 78.1\\  
    %  & \multirow{-2}{*}{MipNeRF-360} & large        & 0.289    & 6.645    & 31.540    & 105.2  & 78.1\\ \cline{2-8} 
     & NeRF-SR     & 0.407    & 9.316    & 29.369   &  0.891 & 129.6  & 46.7\\ \cline{2-8} 
     & Ours & 0.187 & 4.724 & 29.620  & 0.893 & 0.27 & 15.3\\  
\multirow{-7}{*}{Room}  
& Ours-$\mathcal{L}_1$    & 0.304    & 7.732    & 31.147   & 0.912  & 0.27 & 15.3\\ \bottomrule
\end{tabular}
\end{table*}

\begin{table*}[!h]
\centering
\begin{tabular}{l|ccccccc}
\toprule
\multirow{2}{*}{Scene} & \multirow{2}{*}{Method} & \multicolumn{2}{c}{Perception metrics}& \multicolumn{2}{c}{Distortion metrics}   & Inference time      & Cache memory\\ \cline{3-6} 
 &   &  LPIPS $\downarrow$ & NIQE $\downarrow$ & PSNR $\uparrow$ & SSIM $\uparrow$ & (s) $\downarrow$ & (GB) $\downarrow$ \\\midrule
     & Plenoxels  & 0.409    & 8.584    & 24.896  &  0.792 & 2.8    & 30.8\\ 
    %  & \multirow{-2}{*}{Plenoxels}   & large        & 0.410      & 8.647     & 24.926   & 5.4  & 78.7\\ \cline{2-8} 
     & DVGO  & 0.356    & 6.985    & 25.512  & 0.805  & 5.7  & 37.4\\  
    %  & DVGO-${large}$        & 0.315     & 6.543     & 25.764  &  0.818  & 10.2   & 79.1\\  
     & JaxNeRF    & 0.335    & 6.030    & 25.839  & 0.814  & 134.7  & 77.8\\ 
    %  & \multirow{-2}{*}{JaxNeRF}     & large        & 0.316    & 6.092     & 26.147   & 279.3  & 77.8\\ \cline{2-8} 
     & MipNeRF-360  & 0.296    & 5.176    & 25.312  & 0.828  & 51.3   & 78.1\\  
    %  & \multirow{-2}{*}{MipNeRF-360} & large        & 0.274    & 5.033    & 25.122   & 105.2  & 78.1\\ \cline{2-8} 
     & NeRF-SR      & 0.454    & 9.857    & 24.230  & 0.782  & 129.6  & 46.7\\ \cline{2-8} 
     & Ours & 0.193 & 4.672 & 25.121 & 0.796 & 0.26 & 18.0  \\  
\multirow{-7}{*}{T-rex} 
& Ours-$\mathcal{L}_1$     & 0.324     & 6.716     & 26.276  &  0.834 & 0.26 & 18.0  \\ \bottomrule
\end{tabular}
\caption{Detailed results on 4K-LLFF dataset.}
\label{tab:llff}
\end{table*}

\begin{figure*}[p]
\begin{minipage}{\linewidth}
  \centering
  \includegraphics[width=\linewidth]{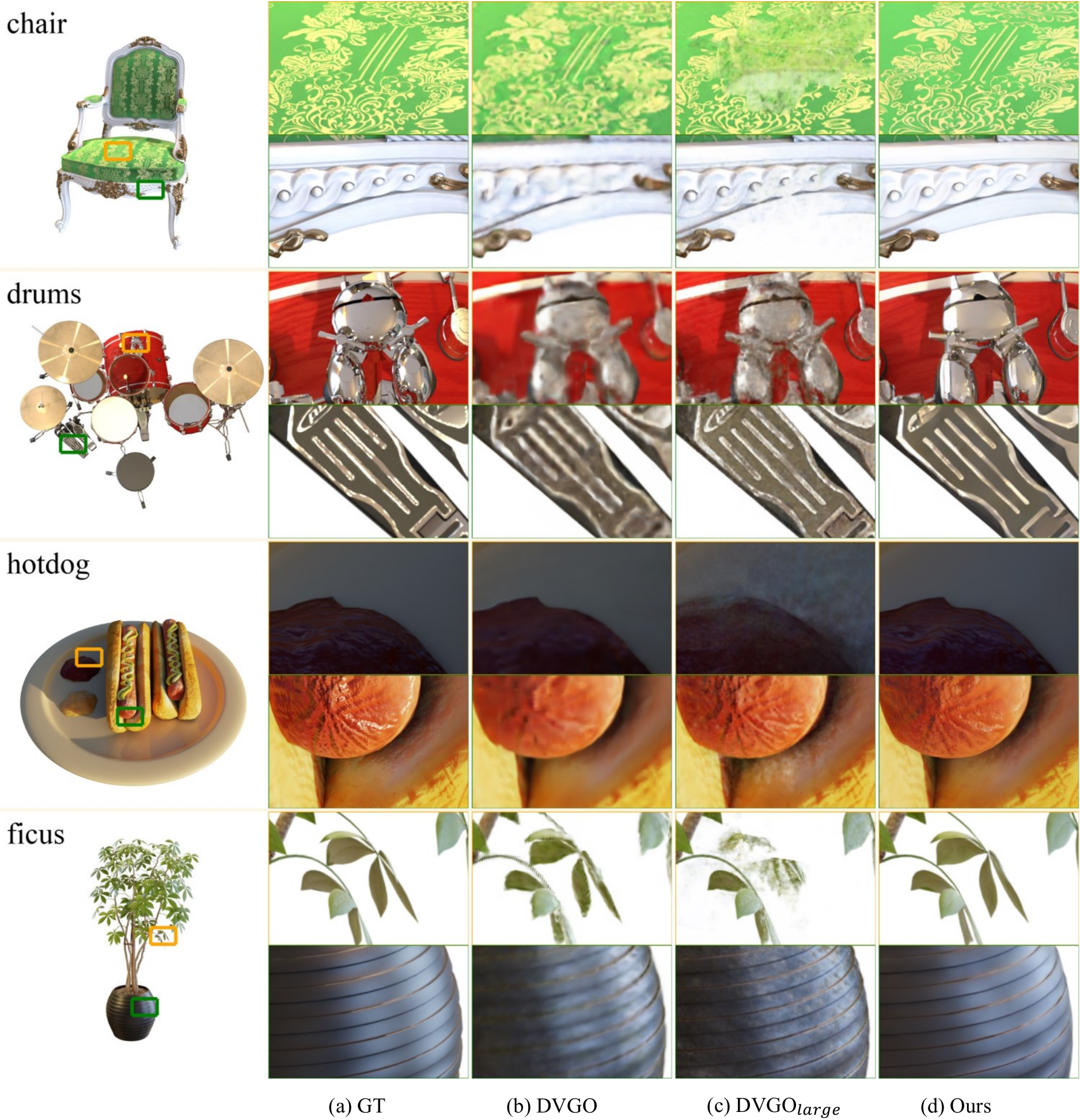}
\end{minipage}
\end{figure*}

% \section{Conclusion}

%%%%%%%%% REFERENCES
% {\small
% \bibliographystyle{ieee_fullname}
% \bibliography{egbib}
% }

% \end{document}

\end{document}